\documentclass[10pt,journal,compsoc]{IEEEtran}

\ifCLASSOPTIONcompsoc
  \usepackage[nocompress]{cite}
\else
  \usepackage{cite}
\fi

\ifCLASSINFOpdf
\else
\fi

\ifCLASSOPTIONcompsoc
 \usepackage[caption=false,font=footnotesize,labelfont=sf,textfont=sf]{subfig}
\else
 \usepackage[caption=false,font=footnotesize]{subfig}
\fi

\usepackage{url}
\usepackage{algorithm}
\usepackage{algorithmic}
\usepackage{amssymb}
\usepackage{amsthm}
\usepackage{bm}
\usepackage{bbm}
\usepackage{booktabs}
\usepackage[switch]{lineno}

\usepackage{amsmath}

\newtheorem{theorem}{Theorem}[section]	

\newtheorem{definition}[theorem]{Definition}
\newtheorem{claim}[theorem]{Claim}

\newtheorem*{rep@theorem}{\rep@title}
\newcommand{\newreptheorem}[2]{%
\newenvironment{rep#1}[1]{%
 \def\rep@title{#2 \ref{##1}}%
 \begin{rep@theorem}}%
 {\end{rep@theorem}}}
\newreptheorem{theorem}{Theorem}
\newreptheorem{lemma}{Lemma}
\newreptheorem{claim}{Claim}

\DeclareMathOperator*{\argmax}{arg\,max}

\usepackage{mathtools}
\usepackage{color}

\usepackage[noabbrev]{cleveref}

\usepackage{etoolbox}
\usepackage{multirow}
\makeatletter
\patchcmd{\@makecaption}
  {\scshape}
  {}
  {}
  {}
\makeatother
\usepackage{color, colortbl}
\usepackage{mathtools}
\usepackage{esvect}

\definecolor{Back}{gray}{0.8}

\begin{document}

\title{
  Temporal Aggregation and Propagation Graph Neural Networks for Dynamic Representation
}

\author{
	Tongya~Zheng,
	Xinchao~Wang,
	Zunlei~Feng,
	Jie~Song,
	Yunzhi~Hao,
	Mingli~Song*,
	Xingen~Wang,
	Xinyu~Wang,
	Chun~Chen
	\IEEEcompsocitemizethanks{
	\IEEEcompsocthanksitem Tongya Zheng is with the Big Graph Center, College of Computer Science of Hangzhou City University, and the College of Computer Science of Zhejiang University, Hangzhou, China. Email: tyzheng@zju.edu.cn.
	\IEEEcompsocthanksitem Xinchao Wang is with the Department of Electrical and Computer Enigneering, National University of Singapore, Singapore. Email: xinchao@nus.edu.sg.
	\IEEEcompsocthanksitem Zunlei Feng and Jie Song are with the School of Software Technology, Zhejiang University, Hangzhou, China. Email: \{zunleifeng, sjie\}@zju.edu.cn.
	\IEEEcompsocthanksitem Yunzhi Hao, Xingen Wang, Xinyu Wang and Chun Chen are with the College of Computer Science, Zhejiang university, Hangzhou, China. Email: \{ericohyz, brooksong, newroot, wangxinyu, chenc\}@zju.edu.cn.\protect
	\IEEEcompsocthanksitem Mingli Song is the corresponding author, and is with the Big Graph Center, College of Computer Science of Hangzhou City University, the ZJU-Bangsun Joint Research Center, and the Shanghai Institute for Advanced Study, Zhejiang University, Hangzhou, China. E-mail: brooksong@zju.edu.cn. \protect\\
	}

}

\markboth{IEEE TRANSACTIONS ON KNOWLEDGE AND DATA ENGINEERING, APRIL 2023}%
{Zheng \MakeLowercase{\textit{et al.}}}

\IEEEtitleabstractindextext{%
\begin{abstract}
	Temporal graphs exhibit dynamic interactions between nodes over continuous time, whose topologies evolve with time elapsing.
	The whole temporal neighborhood of nodes reveals the varying preferences of nodes.
	However, previous works usually generate dynamic representation with limited neighbors for simplicity, which results in both inferior performance and high latency of online inference.
	Therefore, in this paper, we propose a novel method of temporal graph convolution with the whole neighborhood, namely Temporal Aggregation and Propagation Graph Neural Networks (TAP-GNN).
	Specifically, we firstly analyze the computational complexity of the dynamic representation problem by unfolding the temporal graph in a message-passing paradigm.
	The expensive complexity motivates us to design the AP (aggregation and propagation) block, which significantly reduces the repeated computation of historical neighbors.
	The final TAP-GNN supports online inference in the graph stream scenario, which incorporates the temporal information into node embeddings with a temporal activation function and a projection layer besides several AP blocks. 
	Experimental results on various real-life temporal networks show that our proposed TAP-GNN outperforms existing temporal graph methods by a large margin in terms of both predictive performance and online inference latency.
	Our code is available at \url{https://github.com/doujiang-zheng/TAP-GNN}.
\end{abstract}

\begin{IEEEkeywords}
Graph Embedding, Graph Neural Networks, Temporal Graph, Dynamic Graph, Link Prediction
\end{IEEEkeywords}
}

\maketitle

\IEEEdisplaynontitleabstractindextext

\IEEEpeerreviewmaketitle

\IEEEraisesectionheading{\section{Introduction}\label{sec:introduction}}

\begin{figure}[t]
    \centering
    \includegraphics[width=0.45\textwidth]{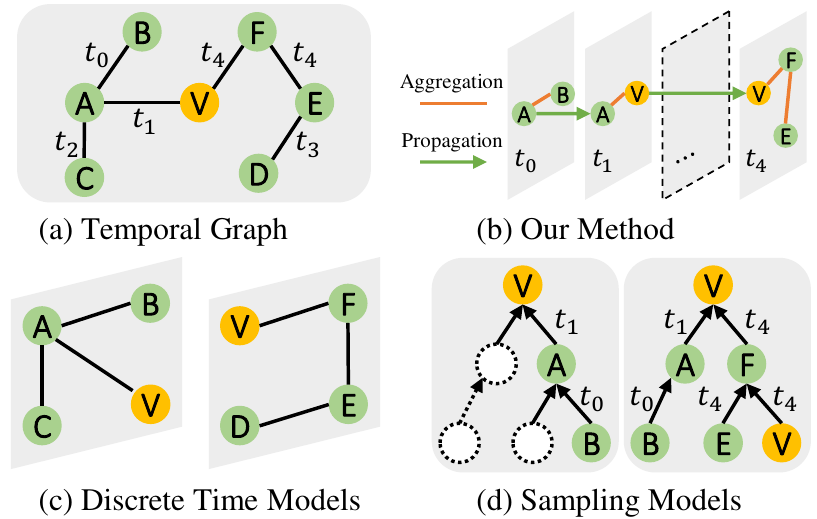}
    \caption{An illustrative example of a temporal graph and the temporal neighborhood of the node $V$ of three compared methods. 
    (a) A temporal graph, also named continuous-time dynamic graph, evolves continuously over time with node occurrences and interactions. 
    (b) Our method aims at accessing the whole varying neighborhood of dynamic nodes, which is realized via the historical nodes. The time points $\{t_2, t_3\}$ which don't involve node $V$ are omitted for clarity here. 
    (c) Discrete-time models slice the continuous-time dynamic graph into a sequence of static graph snapshots, where the temporal information is lost inside the snapshots. 
    (d) Sampling models inspired by \emph{GraphSAGE}~\cite{hamilton2017graphsage} construct a subgraph of the stable topology for each dynamic node.}
    \label{fig:temporal-graph}
\end{figure}

\IEEEPARstart{T}{emporal} interactions construct numerous dynamically variable temporal graphs in many domains, such as bitcoin transaction network, e-commerce purchasing network, movie rating network, social network, etc~\cite{bennett2007netflix,kunegis2013konect,snapnets,network-data}. 
As depicted in Fig.~\ref{fig:temporal-graph}(a), the temporal graph keeps evolving with node occurrences and interactions, with each interaction marked with a timestamp.
In real-life networks, it is crucial to capture node dynamics in temporal graphs to provide accurate predictions on customer preferences and anomalous users~\cite{nguyen2018ctdne,zuo2018htne,tnode,kumar2019jodie,tgat_iclr20,wang2020apan}.
On {the} one hand, recommending items for customers requires taking into account both temporal interests and long-range user preferences~\cite{kumar2019jodie,ying2018pinsage,wang2018billion}.
On the other hand, predicting if a user is malicious in online payment systems defends the system security~\cite{wang2020apan}, and detecting the surge of network traffic guarantees the network service~\cite{network2013zhang}, which {relies} on fast responses to suspicious behaviors.

Learning low-dimensional embeddings for nodes in graphs is a promising method for downstream tasks like node classification (classifying if a user is malicious), link prediction (predicting whether two users will be friends in the future), etc.
The majority of graph embeddings researches~\cite{perozzi2014deepwalk,grover2016node2vec,kipf2016semi,hamilton2017graphsage,embedding2018survey} like graph neural networks (GNNs) focus on static graphs, while are not yet capable of generating dynamic node representation on temporal graphs.
Also, existing industrial practices may suffer from the loss of evolutionary patterns, which mainly involve static graph embedding techniques~\cite{ying2018pinsage,wang2018billion}.

Previous works on temporal graphs can be roughly divided into two categories: discrete-time models as shown in Fig.~\ref{fig:temporal-graph}(c) and sampling models as shown in Fig.~\ref{fig:temporal-graph}(d), which essentially differ in how to model the varying temporal neighborhood of nodes. 
Discrete-time models, such as TNODE~\cite{tnode} and EvolveGCN~\cite{evolvegcn}, usually slice the whole temporal graph into a sequence of graph snapshots, where the temporal information inside a snapshot is lost.
In contrast, sampling models~\cite{tgat_iclr20,wang2020apan} construct the k-hop temporal subgraphs for each dynamic node (i.e., a node with a query timestamp) and conduct message-passing on the subgraphs.
The neighborhood of sampling models is restricted {by} the predefined window despite its flexibility.

In summary, two challenges {remain} unsolved for existing methods of temporal graphs.
The first challenge is how to capture the whole temporal neighborhood of nodes with time elapsing.
If considering the whole temporal neighborhood, discrete-time models will turn to static graph models; sampling models will suffer from huge complexity (hundreds of times of complexity compared to the original temporal graph) even with 1-hop neighbors.
The second challenge is how to perform online inference when the interactions between nodes arrive at the finest temporal granularity (e.g., every second or millisecond).
The inference latency of most methods is positively related to the neighborhood size.
Notably, the neighborhood of sampling models increases exponentially with the number of layers.
It requires a trade-off between the representation quality and the inference latency.

\textbf{Present work.} 
Therefore, in this paper, we present Temporal Aggregation and Propagation Graph Neural Networks (TAP-GNN), which generates dynamic node embeddings with the whole temporal neighborhood and supports online inference in the graph-stream scenario.

Firstly, the temporal graph is unfolded into a message-passing graph as shown in Figure~\ref{fig:temporal-graph}(b) without the propagation links, where the size of temporal nodes is linear with the size of edges.
The temporal nodes are densely connected with their temporal neighbors following the message-passing paradigm, resulting in huge complexity.
Further, a decomposition block, namely aggregation and propagation (AP), is proposed to reduce the repetitive computation in the proposed message-passing graph, as illustrated in Fig.~\ref{fig:temporal-graph}(b).
The additional propagation links send the aggregated neighbor information in the past to nodes in the future, which is mathematically equal to existing graph convolution kernels.
The AP block both inherits the merits of GNNs and shows linear complexity with the number of temporal edges.

The final learning framework, Temporal Aggregation and Propagation Graph Neural Networks (TAP-GNN) aims at generating node embeddings at any time with various temporal information. 
It incorporates the time point encodings with a temporal activation function and a projection layer.

Moreover, TAP-GNN supports online inference due to the decomposition property of the AP block, whose complexity is only related to the size of upcoming new interactions.

To evaluate our framework, we conduct experiments over various real-life temporal graphs on two prediction tasks: future link prediction and dynamic node classification.
In the experiments, our proposed TAP-GNN demonstrates robust and better performance than the state-of-the-art temporal graph methods. 
It implies that the TAP-GNN benefits significantly from the \emph{whole temporal neighborhood} of nodes.
Meanwhile, the inference latency of TAP-GNN is only linear with the number of layers, compared with the exponential increasing time cost of TGAT~\cite{tgat_iclr20}.

Our contributions can be summarized as follows:
\begin{itemize}
    \item The dynamic node representation problem on temporal graphs is formulated in a message-passing paradigm, and its computational complexity is analyzed theoretically and empirically.
    \item An efficient AP block is proposed to perform graph convolutions on the whole temporal neighborhood of nodes, whose computational complexity is only linearly dependent on the number of temporal edges.
    \item The TAP-GNN framework provides dynamic node representation and supports online inference in the graph stream scenario built on top of the proposed AP block.
    \item Experimental results show that our proposed TAP-GNN outperforms existing temporal graph methods in terms of both predictive performance and online inference latency.
\end{itemize}

The rest of our paper is organized as follows.
Section 2 reviews the related works.
In Section 3, we introduce the message-passing temporal graphs, the proposed AP block, and the TAP-GNN framework.
The computational complexities of each proposed module are discussed separately.
The experimental results and ablation studies are discussed in Section 4.
Section 5 concludes the paper.

\section{Related Works}

\subsection{Static Graph Embedding}

Decades ago, the emerging big graph data on the website had already attracted much concern of researchers~\cite{page1999pagerank,roweis2000nonlinear,belkin2002laplacian,bennett2007netflix,sun2011pathsim}.
The learned low-dimensional embeddings for nodes, edges, and subgraphs can be easily integrated with the machine learning algorithms like Logistic Regression, Random Forest, and Gradient Boosting Decision Trees to perform risk monitoring~\cite{network2013zhang}, link prediction~\cite{sun2011pathsim}, item recommendation~\cite{bennett2007netflix}, etc.
Recently, researchers interested in graph mining {have been} motivated by the great success of deep learning methods in computer vision~\cite{alexnet} and natural language processing~\cite{word2vec}.
The deep graph embedding methods can also be categorized into two kinds: graph convolutional networks~\cite{gnn2020survey} and skip-gram models~\cite{embedding2018survey}.

On {the} one hand, graph neural networks~\cite{deffe2016gcn} are initially defined in the spectral space, which is inspired by the convolution operations defined on the image grid. 
GCN~\cite{kipf2016semi} simplifies the learning framework of~\cite{deffe2016gcn} and shows a successful application on semi-supervised node classification tasks.
GraphSAGE~\cite{hamilton2017graphsage} proposes an inductive learning paradigm on large-scale graphs, which samples subgraphs for each node.
Meanwhile, GAT~\cite{velivckovic2017gat} introduces a learnable attention mechanism to impose different importances over neighbors.
The attention mechanism not only achieves better performance on several benchmarks but also presents an explainable result by the neighbor importances.

On the other hand, DeepWalk~\cite{perozzi2014deepwalk} is the pioneering work to introduce the skip-gram models~\cite{word2vec} into graph representation learning with the sampled node sequences by random walks.
Node2Vec~\cite{grover2016node2vec} design two hyper-parameters to control the random walk preferences, while LINE~\cite{tang2015line} {explores} the high-order relations in the graph topology.
Their shortcomings are obvious: the generated embeddings are not task-oriented, and a new model is required to deal with the additional side information.
The interested readers can refer to the survey papers~\cite{embedding2018survey,gnn2020survey} for an overview of graph embedding methods in general.

Moreover, these methods have successful applications in industrial scenarios such as commodity recommendation and risk monitoring~\cite{ying2018pinsage,wang2018billion,zhou2019dien,wang2019ngcf,kumar2019jodie}.
However, these static graph embedding methods are not designed to capture node dynamics with the continuously emerging node interactions.
Hence, an additional model is needed to perform real-time predictions, which may lose the evolutionary pattern in temporal graphs.

\subsection{Discrete-Time Dynamic Graph Embedding}

As illustrated in Fig.~\ref{fig:temporal-graph}(c), discrete-time dynamic graph (DTDG) embedding methods slice the whole temporal graph into a sequence of graph snapshots.
The predefined time window choice is a crucial part of DTDG methods, which causes high {costs} for hyper-parameter search.
Intuitively, the smaller time window models the real scenarios more precisely with the increasing computation cost.
As a result, DTDG methods are designed for scenarios that only need to update embeddings periodically.

DTDG methods mainly contain two parts: the first part is to capture the evolutionary patterns; the second part is to fuse the historical information with current embeddings. 
For the first part, DynamicTriad~\cite{zhou2018dynamic} models the probability that open triads evolve into closed triads; SPNN~\cite{subgraph2018meng} studies high-order dependencies of the induced subgraphs; EPNE~\cite{epne2020wang} incorporates both periodic and non-periodic patterns into the learned embeddings.
For the second part, Know-Evolve~\cite{trivedi2017know}, E-LSTM-D~\cite{elstm2019chen}, and TNODE~\cite{tnode} use an LSTM to fuse the historical information and current information.
In addition, EvolveGCN~\cite{evolvegcn} learns to generate the transformation matrix for each separate graph snapshot in a recurrent paradigm.
Nevertheless, DTDG methods cannot generate dynamic node embeddings to meet the requirements of continuous-time dynamic graphs.

\subsection{Continuous-Time Dynamic Graph Embedding}

Continuous-time dynamic graph (CTDG), which we use with temporal graph exchangeably, keeps evolving with the appearance, disappearance, and changing of edges and nodes~\cite{nguyen2018ctdne}.
Recent advances for dynamic node representation on temporal graphs have been summarized in~\cite{survey2020}.
While they focus on the encoder-decoder techniques, we mainly review existing solutions to the aforementioned challenges in this paper.

Methods tackling the varying neighborhood challenge are mostly recurrent models.
HTNE~\cite{zuo2018htne} adopts the Hawkes process and an attention mechanism to model the influence of temporal neighbors; 
DyRep~\cite{trivedi2018dyrep} discriminates the topological evolution from node interactions and introduces a two-time scale temporal point process model;
JODIE~\cite{kumar2019jodie} uses two mutually recursive RNNs for users and items separately and proposes a projection operation to predict the node trajectory; 
TigeCMN~\cite{zhang2020tigecmn} adopts a key-value memory network to encode the network dynamics with sequentially {updated} operations.
Differently, CTDNE~\cite{nguyen2018ctdne} incorporates temporal random walks into the static skip-gram models but can only obtain the final state for each node.
The aforementioned CTDG methods {mainly} rely on the local node embedding update with the upcoming interactions.
Consequently, the influence of historical neighbors in these models will gradually vanish, and high-order neighbors are neglected by these models, which have been proved useful in static GNNs~\cite{ying2018pinsage,wang2019ngcf}.
Our proposed TAP-GNN treats the historical neighbors according to their contributions to the generated embeddings and can be easily extended to obtain the high-order temporal neighbors.

In contrast, CTDG methods based on GNNs are naturally designed for the high-order neighborhood on temporal graphs but cannot capture the whole neighborhood due to the neighbor dropout technique.
Hence,  previous works on GNNs can only partially inject temporal information and perform online inference.
TGAT~\cite{tgat_iclr20} and APAN~\cite{wang2020apan}, which are the most related works to our paper, construct message-passing subgraphs for each dynamic node inspired by GraphSAGE~\cite{hamilton2017graphsage}.
Detailedly, TGAT~\cite{tgat_iclr20} proposes a novel temporal encoding kernel and implements an attention-based graph convolution model.
However, TGAT suffers from the exponentially expanding neighborhood and the information loss of the unseen neighbors.
APAN makes substantial improvements over TGAT in consideration of online inference latency but also inherits the restricted neighbor size.
Therefore, they both are unable to access the information from the whole temporal neighbors.

\section{Methods}

In this section, we present the message-passing paradigm on temporal graphs for dynamic node representation. 
An alternative message-passing temporal graph (MPTG) is constructed upon the definitions of temporal nodes and neighbors. 
However, further analysis shows the cumbersome computational complexity of the MPTG. 
An efficient Aggregation-Propagation (AP) block is proposed to reduce the repetitive computations in the MPTG, which is mathematically equivalent to the message-passing graph convolutions. 
The final learning framework, Temporal Aggregation-Propagation Graph Neural Networks (TAP-GNN) consists of a sequence of AP blocks and a projection layer, aiming at generating dynamic node embeddings. 
Moreover, the TAP-GNN can be easily extended to support online inference, whose time complexity is linearly dependent on the size of upcoming new interactions.

\subsection{Dynamic Node Representation Formulation}

A temporal graph can be defined as $G = (V, E, \mathcal{T})$, where $E$ is extended as a set of temporal edges, and $\mathcal{T}:E\rightarrow \mathbb{R}^+$ is a function that maps each edge to a corresponding timestamp\cite{nguyen2018ctdne}. A common assumption is that edges in temporal graphs are formed according to some underlying distribution~\cite{perozzi2014deepwalk,nguyen2018ctdne,yang2020understanding}.
The neighbors of node $v$ at time $t$ can be drawn from $u \sim p(\cdot | v, t)$ in temporal graphs. In general, our target of graph embedding is to learn an embedding function revealing the underlying data distribution. Let $h(\cdot, t)$ {be} the dynamic node embedding function, {and} the dot product be the similarity function. The dynamic node representation problem can be written as
\begin{equation}
  \argmax_{h} p(u | v, t) = \frac{\exp (h(u, t) \cdot h(v, t)) }{ \sum_{u' \in V} \exp (h(u', t) \cdot h(v, t)) },
  \label{eq:mle}
\end{equation}
where $(u, v, t)$ is an observed interaction, $t$ is the time point, and $u'$ iterates all nodes of $V$. In practice, the negative sampling loss~\cite{word2vec,yang2020understanding} is usually adopted instead as Eq.~\ref{eq:mle} to accelerate the training speed.

\subsection{{Preliminary of Graph Neural Networks}}

{
Existing Graph Neural Networks (GNNs) mainly follow the message-passing paradigm that aggregates neighborhood embeddings recursively to obtain high-order neighborhood information. 
Formally, the graph convolution operation is defined as
\begin{equation}
    h(v) = g^\star(\{h(u), \forall u \in \mathcal{N}(v) \}),
    \label{eq:gcn}
\end{equation}
in~\cite{kipf2016semi,hamilton2017graphsage}, where $h(\cdot)$ is an embedding function, $\mathcal{N}(\cdot)$ is the neighborhood function, $g^\star(\cdot)$ is the graph convolution kernel.
The detailed $g^\star(\cdot)$ kernels can be listed as follows:
\begin{equation}
	\begin{split}
		\textbf{GCN: } & \sum_{u \in \mathcal{N}(v)} \frac{W h_u}{\sqrt{\deg(u)} \sqrt{\deg(v)}}, \\
		\textbf{MEAN: } & \sigma(\text{mean}\{ W h_u + b, \forall u \in \mathcal{N}(v)\}\}), \\
		\textbf{POOL: } & \sigma(\text{max}\{ W h_u + b, \forall u \in \mathcal{N}(v) \}), \\
	\end{split}
	\label{eq:graph-kernel}
\end{equation}
where $h(u)$ is written as $h_u$, $\deg(\cdot)$ denotes the degree of nodes, $\sigma(\cdot)$ is a non-linear activation function, $W$ is a transformation matrix, mean and max are reduction operations.
The attention kernel is provided in Appendix~\ref{appendix:attention}.
}

\subsection{Message-Passing Temporal Graph}
\label{sec:mptg}
Although $h(v, t)$ requires dynamic representation for node $v$ with the varying $t$, the representation should be related {to} the limited node interactions. The observation motivates us to define the temporal nodes restricted to the observed time points. Representation for the infinite time dimension can be solved with an additional projection layer. Let $T_K=\{t_k | (u, v, t_k) \in  E, u \in V, v \in  V, t_k \in \mathbb{R}^+\}$ denotes the set of observed time points. For simplicity, we model the node dynamics with the observed $T_K$ as:
\begin{definition}
	Temporal nodes $V^T=\{v_{t_k} | (u, v, t_k) \in E \}$ associates each node with its occurrences in temporal edges.
\end{definition}

\begin{definition}
	\label{def:temporal-neighbor}
	Temporal neighbors is defined with a dynamic query $(v,t)$ as $\mathcal{TN}(v, t) = \{u_{t_l} | (u, v, t_l) \in E, t_l \le t, u_{t_l} \in V^T\}$, where $v \in V, t \in \mathbb{R}^+$. Specifically, the temporal neighbors of $v_{t_k}$ is defined as $\mathcal{TN}(v, t_k)$, which is abbreviated as $\mathcal{TN}(v_{t_k})$ for simplicity.
\end{definition}

For temporal nodes, it's worth noting that the set of temporal nodes $V^T$ is not the cartesian product of nodes $V$ and $T_K$ since $v_{t_l} \notin V^T$ if there are no interactions of $v$ at $t_l$. Nodes are only connected with their active time points, as shown in Fig.~\ref{fig:temporal-graph}(b). Since temporal graphs may have parallel and dynamic edges between nodes, the spectral graph convolution way~\cite{kipf2016semi} is not feasible in our settings.

Generally speaking, the definition of temporal neighbors is suitable for any dynamic node $(v, t)$ even if $t$ never appears in the observed $T_K$. It means that we can generate \textbf{dynamic node representation} for any $v_t$ as long as node embeddings on $V^T$ are obtained. Moreover, the defined temporal neighbors are associated with each other via directional links, i.e. $u_{t_l} \in \mathcal{TN}(v_{t_k})$ doesn't imply $v_{t_k} \in \mathcal{TN}(u_{t_l})$ except $t_l = t_k$. A simple corollary is that temporal neighbors of one node are also temporal neighbors of it in the future, i.e., $\mathcal{TN}(v, t_j) \subseteq \mathcal{TN}(v, t_i)$ when $t_j \le t_i$.

So far, we can build the message-passing temporal graph (MPTG) as $G_{MP}=(V^T, E^T)$, where $V^T$ is the set of temporal nodes, and $E^T = \{(u_{t_l}, v_{t_k}) | u_{t_l} \in \mathcal{TN}(v_{t_k}), v_{t_k} \in V^T \}$ is the set of directional links between temporal nodes. The messages of the temporal neighbors can thus be sent to the destination nodes on the $G_{MP}$, i.e., message passing operations on temporal graphs. High-order temporal neighbors can be accessed by stacking layers of the $G_{MP}$ as described in~\cite{kipf2016semi,hamilton2017graphsage,message-passing}.
{
For instance, if we focus on V and its one-hop neighbor \{A,F\} in Fig.~\ref{fig:temporal-graph}(d), we will get three temporal nodes $V^T=\{\text{V}_{t_1},\text{V}_{t_4},\text{A}_{t_1},\text{F}_{t_4}\}$.
At the time $t_1$, the edge set $E^T$ contains one temporal edge $\text{V}_{t_1}\leftarrow \text{A}_{t_1}$.
At the time $t_4$, the edge set $E^T$ contains two temporal edges $\{ \text{V}_{t_4} \leftarrow \text{A}_{t_1}, \text{V}_{t_4} \leftarrow \text{F}_{t_4}\}$, where $\text{V}_{t_4}$ shares a same edge with $\text{V}_{t_1}$ at $t_1$.
The overall $E^T$ is the union set of the edge set at each time point, namely $\{\text{V}_{t_1}\leftarrow \text{A}_{t_1}, \text{V}_{t_4} \leftarrow \text{A}_{t_1}, \text{V}_{t_4} \leftarrow \text{F}_{t_4}\}$.
}
\begin{table}[!t]
	\caption{The statistical size of $E$ and $E^T$ on different temporal graphs. $\vert E \vert$ is the number of edges in the dataset. $\vert E^T \vert$ is the number of directional links between temporal nodes. {The ratio} refers to the proportion of $\vert E^T \vert$ and $\vert E \vert$. Here thousand, million, and billion are abbreviated as K, M, and B.}
	\label{tab:graph-size}
	\centering
	\resizebox{0.4\textwidth}{!}{

	\begin{tabular}{rrrr}
		\toprule
		Temporal Graph & $\vert E \vert$ & $\vert E^T \vert$ & Ratio  		\\
		\midrule

		fb-forum   & 33K & 9.9M & 296 		\\
		soc-sign-bitcoinotc & 35K & 4.7M & 134  \\
		ia-escorts-dynamic  & 50K & 2.6M & 51 \\

		ia-movielens-10m   & 95K & 64M & 671  \\
		soc-wiki-elec      & 107K & 17M & 164  \\
		ia-slashdot-reply  & 140K & 13M & 94  \\

		Wikipedia		& 167K & 39M & 253	\\

		Reddit			& 672K & 5.7B & 8537	\\

		\bottomrule
	\end{tabular}
	}

\end{table}

However, the alternative MPTG based on the temporal graph may incur the computation complexity problem because of the repeated links with temporal neighbors. We provide an analysis of the cardinality of temporal nodes $V^T$ and directional links $E^T$ as {follows}.
\begin{claim}
	The size of temporal nodes $\vert V^T \vert$ is $\mathcal{O}(\vert E \vert)$, where $E$ represents the edges in the original temporal graph.
	\label{eq:node-size}
\end{claim}

\begin{claim}
	The size of directional links $\vert E^T \vert$ defined in $G_{MP}$ is $\mathcal{O}(\sum_{v \in V} d_v^2)$, where $V$ is the original node set and $d_v$ denotes the degree of node $v$.
	\label{eq:neighbor-size}
\end{claim}

Claim~\ref{eq:node-size} implies that the size of temporal nodes increases linearly with the dynamic interactions on temporal graphs. A new temporal node is created {when} each new interaction {is} marked with a unique timestamp.

Claim~\ref{eq:neighbor-size} indicates that directional links expand rapidly with the increasing dynamic interactions. Table~\ref{tab:graph-size} also shows the surprising result on experimental datasets, where ratios of small graphs still exceed $50$, while the largest expansion ratio even reaches $8,537$. The resulting message-passing graph $G_{MP}=(V^T, E^T)$ may not fit the main memory for medium graphs. Therefore, we have to design a lightweight mechanism to replace the burdensome $G_{MP}$.

\begin{figure}[t]
	\centering
	\includegraphics[width=0.48\textwidth]{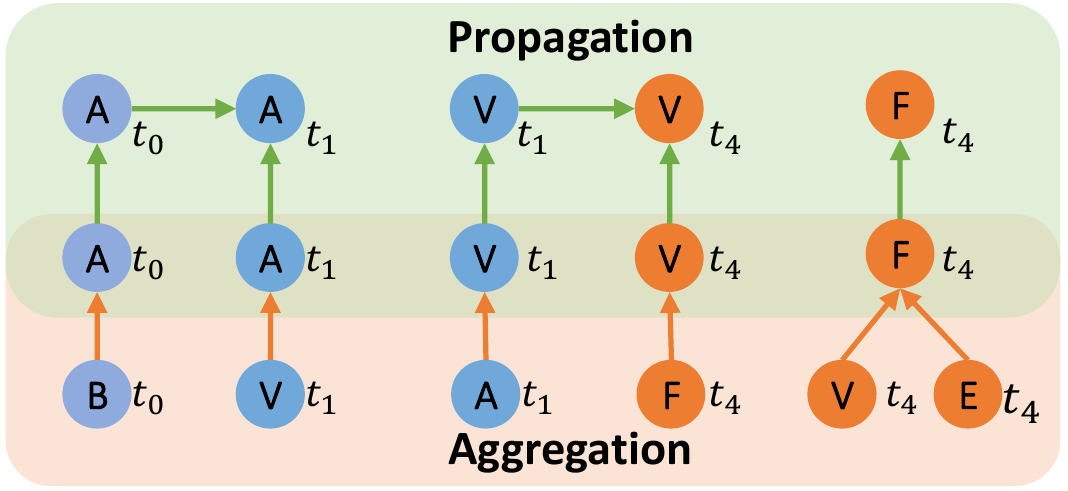}%
	\caption{An illustrative example of the proposed Aggregation-Propagation block. The light orange module denotes the aggregation layer, and the light green module denotes the propagation layer. Nodes of the same color are active at the same time point.}
	\label{fig:agg-prop-block}
\end{figure}

\subsection{Aggregation-Propagation Block}
\label{section:ap-block}

The proposed block comes from an observation that historical node embeddings already contain the information of historical neighbors. Let's take a single node $v$ and its consecutive active time points $t_i, t_{j} (i > j)$ into account. Here $t_{i}$ and $t_{j}$ only represent the relative sequential order. Referring to the Definition~\ref{def:temporal-neighbor} $\mathcal{TN}(\cdot)$, $\mathcal{TN}(v_{t_j})$ is a subset of $\mathcal{TN}(v_{t_i})$. Hence, the difference set of the two sets is $\{ u_{t_{i}} | (u, v, t_{i}) \in E \}$, which is only related with $t_{i}$. Let $\mathcal{DN}(v_{t_i})$ denote the difference set, then $\mathcal{TN}(v_{t_{i}}) = \mathcal{TN}(v_{t_{j}}) \cup \mathcal{DN}(v_{t_{i}})$ and $\mathcal{TN}(v_{t_j}) \cap \mathcal{DN}(v_{t_i}) = \emptyset$. It indicates that the aforementioned temporal graph convolution can be decomposed sequentially as long as it doesn't involve \emph{commutative computations} between $\mathcal{TN}(t_j)$ and $\mathcal{DN}(v_{t_{i}})$. 
Fortunately, we find that the commonly adopted graph convolution kernels on static graphs fit this property.

Thus, the Aggregation-Propagation (AP) block, as depicted in Fig.~\ref{fig:agg-prop-block}, is designed to save the repeated computation in $G_{MP}$.
We illustrate the overall workflow of the proposed AP block before going into the technical decomposition details of several graph convolution methods.
The AP block consists of two sequential layers, termed aggregation layer and propagation layer, {which refer to AGG and PROP used in Algo.~\ref{alg:tgl}, respectively.}
The aggregation layer is a lightweight message-passing graph and written as $G_{\text{AGG}} = (V^T, {O^T})$, where ${O^T = \{ (u_{t_i}, v_{t_i}) | (u, v, t_i) \in E \}}$ connects the temporal nodes only at the same time point.
It works in a message-passing way that sends the neighbor information to the temporal nodes like $A_{t_1} \rightarrow V_{t_1}; F_{t_4} \rightarrow V_{t_4}$.
The propagation layer reuses the historical node embeddings to compute the temporal node embeddings and works with a linked list $L^T = \{ \text{sorted}(\{ v_{t_i} | v_{t_i} \in V^T \}),  \forall v \in V \}$, which organizes the temporal nodes $v_{t_i}$ referring to the same origin node $v$ into a sorted list.
It accumulates the node embeddings chronologically as $V_{t_1} \rightarrow V_{t_4}$.

{
Let $\textbf{H}(V^T)$ be the embeddings of temporal nodes.
Formally, we combine AGG and PROP for a node $v$ to obtain the same embeddings as graph kernels in Eq.~\ref{eq:graph-kernel} regardless of time points, written as
\begin{equation}
	g^\star(\textbf{H}(V^T)) \equiv \text{PROP}(L^T, \text{AGG}(O^T, \textbf{H}(V^T))), 
\end{equation}
where AGG aggregates neighborhood embeddings and PROP summarizes historical embeddings.
Let $v_{t_i}, v_{t_j} (i > j)$ be the consecutive active time points of node $v$, the AGG operation is written as.
\begin{equation}
	a(v_{t_i}) = \text{AGG}(\{h_u, \forall u \in \mathcal{DN}(v_{t_i})\}),
\end{equation}
where $h_u$ is the node embedding of the previous layer, and $\mathcal{DN}(\cdot)$ refers to the neighbor set of the current time point.
The PROP operation is then written as
\begin{equation}
	h(v_{t_i}) = \text{PROP}(a(v_{t_i}), h(v_{t_j})),
\end{equation}
where $h(v_{t_j})$ is $v_{t_j}$'s embedding of PROP, and \text{PROP} performs on  $v_{t_j}, v_{t_i}$ sequentially.

Suppose that $u \in \mathcal{DN}(v_{t_i})$, the decomposition methods are listed as follows, where the attention kernel is provided in Appendix~B.
}

\textbf{GCN}. Aggregation: Let $a(v_{t_i})= \sum_u \frac{h_u}{\sqrt{\deg(u)}}$. Propagation: $h(v_{t_i})=\frac{a(v_{t_i}) + h(v_{t_j}) \cdot \sqrt{\deg(v_{t_j})}} {\sqrt{\deg(v_{t_i})}}$, where $\deg(\cdot)$ denotes the indegree of nodes.

\textbf{MEAN}.
{
Aggregation: $a(v_{t_i}) = \sum_u h_u$. Propagation: $h(v_{t_i}) = \frac{a(v_{t_i}) + h(v_{t_j}) \cdot \deg(v_{t_j})}{\deg(v_{t_i})}$.
}

\textbf{POOL}. Aggregation: $a(v_{t_i}) = \text{max} \{ h_u \}$. Propagation: $h(v_{t_i}) = \text{max}({a(v_{t_i}), h(v_{t_j}}))$.

\begin{figure*}[t]
	\centering
	\includegraphics[width=.8\textwidth, height=0.2\textwidth]{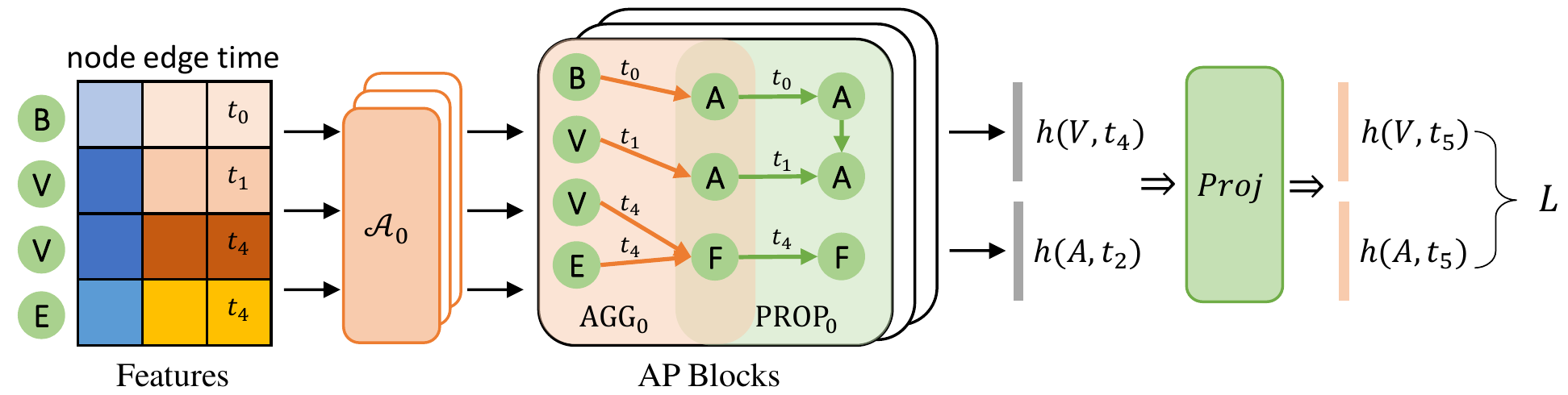}
	\caption{TAP-GNN framework consists of several AP blocks and a projection layer. 
	The temporal activation function $\mathcal{A}$ fuses node features, edge features, and time encodings for temporal nodes before each AP block.
	The AP blocks obtain the temporal node embeddings with the information from high-order temporal neighbors. 
	The projection layer is designed to generate dynamic node representation with respect to the prediction time point. 
	The learning task is the self-supervised future link prediction.}
	\label{fig:framework}
\end{figure*}

\textbf{Complexity analysis.}
Let $N$ be the size of nodes, and $M$ be the size of temporal edges.
The AP block shares the same node set $V^T$ with $G_{MP}$ and preserves the links between nodes at the same time point, whose size is $M$.
Suppose each node has $d$-dimension embeddings, the space complexity of the AP block is $\mathcal{O}(Md)$ since the size of $V^T$ is $M$.
The time complexity of the aggregation layer is $\mathcal{O}(Md)$ because the size of the preserved links is $M$.
The time complexity of the propagation layer is also $\mathcal{O}(Md)$ because it propagates embeddings over temporal nodes.
But it requires more time than the aggregation layer in practice since the propagation cannot be fully paralleled.
Overall, the time complexity of the AP block is $\mathcal{O}(Md)$, where the feature transformation operations are omitted here.

\begin{algorithm}[t]
	\caption{Dynamic Node Representation}
	\label{alg:tgl}
	\begin{flushleft}
		\textbf{Input}: Graph $G(V,E,\mathcal{T})$; Input features $\{x_v, \forall v \in V\}$;
		\newline A batch of queries $(v, t)$;
		\newline Temporal activation functions $\mathcal{A}_k(\cdot, \cdot), \forall k \in \{1,\cdots,K\}$;
		\newline Aggregation functions $\text{AGG}_k, \forall k \in \{1,\cdots,K\}$;
		\newline Propagation functions $\text{PROP}_k, \forall k \in \{1,\cdots,K\}$;
		\newline Projection layer $Proj(\cdot, \cdot)$; 
		\newline
		\textbf{Output}: Dynamic node embeddings $h(v, t)$.
	\end{flushleft}
	\begin{algorithmic}[1]
		\STATE Construct $V^T = \{v_{t_i} | (u, v, t_i) \in E\}$.
		\STATE {Construct $O^T = \{(v_{t_i}, u_{t_i}) | (u, v, t_i) \in E \}$.}
		\STATE Construct $L^T = \{ \text{sorted}(\{v_{t_i} | v_{t_i} \in V^T \}), \forall v \in V\}$.
		\STATE $\textbf{H}_{0}(V^T) \leftarrow \{ x_v, \forall v_{t_i} \in V^T \}$.
		\FOR {$k=1 \cdots K$}
			\STATE $\textbf{H}_{k-1}(V^T) \leftarrow \{ \mathcal{A}_k(\textbf{H}_{k-1}(v_{t_i}), t_i), \forall v_{t_i} \in V^T \}$;
			\STATE {$\textbf{A}_{k}(V^T) \leftarrow \text{AGG}_k(O^T, \textbf{H}_{k-1}(V^T))$;}
			\STATE $ \textbf{H}_{k}(V^T) \leftarrow \text{PROP}_k(L^T, \textbf{A}_{k}(V^T))$;
		\ENDFOR
		\STATE {$h(v, t) \leftarrow \{Proj(\textbf{H}_K(v_{t_n^-}), t_n), \forall (v_n, t_n) \in (v,t)\}$.}

	\end{algorithmic}
\end{algorithm}

\subsection{Temporal Aggregation and Propagation Graph Neural Networks}

The whole process is listed in Algorithm~\ref{alg:tgl}. 
The input includes a temporal graph $G(V, E, \mathcal{T})$, features for all nodes $x_v, \forall v \in V$, and a batch of queries $(v, t)$ asking for dynamic node representation. 

The additional edge features are concatenated with node features for input, as illustrated in Fig.~\ref{fig:framework}.
The learning backbone consists of $K$-layer AP blocks and a final projection layer, where $K$ denotes the iterations of temporal graph aggregation-propagation. 
{Inspired} by~\cite{kumar2019jodie,tgat_iclr20}, a temporal activation function $\mathcal{A}$ is applied before each AP block to inject the temporal information into the node embeddings as
\begin{equation}
	\mathcal{A}(h_v, t_i) = \text{CONCAT}(h_v, \cos(Wt_i + b)),
	\label{eq:activation}
\end{equation}
where $h_v$ is the hidden representation of node $v$, $t_i$ is the corresponding time point, $W t_i + b$ encodes the temporal information, and $\cos(\cdot)$ may capture the repeated patterns of interactions. 
The temporal activation function provides distinct embeddings even {though} some node has few interactions. 
The AP block described in Fig.~\ref{fig:agg-prop-block} works with the help of the proposed temporal nodes $V^T$, a new set of temporal edges ${O^T}$, and a propagation linked list $L^T$. $V^T$ splits nodes at different time points as different temporal nodes, ${O^T}$ links the temporal nodes at the same time point with respect to the original edges, and $L^T$ organizes the temporal nodes $v_{t_i}$ referring to the same node $v$ into sorted lists. 
The temporal node features $h^0(v_{t_i})$ are initialized from the given node features $x_v$. 
At each inner loop of the iterations, temporal activation function $\mathcal{A}_k(\cdot, \cdot)$ injects the temporal information into node features from the previous layer $h^{k-1}(v_{t_i})$. 
The function $\text{AGG}_k$ then aggregates neighbor features $\textbf{H}(V^T)$ into the intermediate node embeddings $\textbf{A}(V^T)$ following the message-passing paradigm~\cite{message-passing}. 
Further, the function $\text{PROP}_k$ propagates the $\textbf{A}(V^T)$ along the sorted node list $L^T$ for each node $v \in V$ independently. 
We can obtain the high-order temporal node embeddings $h^K(v_{t_i})$ after $K$ iterations. 
However, the required dynamic node representation is usually defined upon a future time point on tasks like future link prediction. 
Let $v_{t_n^-}$ be  $\argmax_{t_i} \{v_{t_i} | v_{t_i} \in V^T, t_i < t \}$, which is the latest node embedding for a query $(v_n, t_n)$. 
A projection layer is then used to provide dynamic node representation, which works as
\begin{equation}
	{Proj(h(v_{t_n^-}), t_n) = \text{MLP}(\mathcal{A}(h(v_{t_n^-}), t_n - t_n^-)),}
	\label{eq:proj}
\end{equation}
where MLP denotes a multilayer perceptron, and $\mathcal{A}(\cdot, \cdot)$ is the temporal activation function.

\textbf{Node grouping.} 
The training speed can be boosted much faster with GPUs since \cite{alexnet}. 
Thanks to \cite{dgl2019wang}, the computation speed of graphs can be largely accelerated by grouping nodes with the same degree into a computation bucket.
{Detailed description is provided in Appendix~\ref{appendix:node-grouping}.}

\textbf{Model optimization.} 
We perform future link prediction for the TAP-GNN learning framework in a self-supervised paradigm.
The training objective $L$ is defined as
\begin{equation}
	\begin{split}
	L &= - \sum_{(u, v, t) \in E} \log (\sigma(h(u, t)^\intercal h(v, t))) \\ 
			& - k \cdot \mathbb{E}_{u_n \sim P_n(u)} \log(\sigma(-h(u_n, t)^\intercal h(v, t))),
	\end{split}
	\label{eq:negative-sampling}
\end{equation}
where the loss iterates over the temporal edges, $\sigma(\cdot)$ is the sigmoid function, $k$ denotes the number of negative samples, and $P_n(v)$ denotes the negative sampling distribution on nodes.

\textbf{Complexity analysis.}
The dimensions for all input features and hidden features are assumed $d$ for simplicity.
Let $N$ be the size of nodes, and $M$ be the size of temporal edges.
Each AP block contains a temporal activation function and an aggregation function, so it has $\mathcal{O}(d^2)$ parameters.
Besides, the projection layer has $\mathcal{O}(d^2 + ld^2)$ parameters, where the first part refers to the temporal activation function, the second part refers to the multilayer perceptron, and $l$ denotes the number of layers.
The total parameters of the TAP-GNN is $\mathcal{O}(Kd^2 + ld^2)$, where $K$ means the number of AP blocks.
However, the training of the TAP-GNN requires computing the whole temporal neighborhood.
As a result, the space complexity of the TAP-GNN is $\mathcal{O}(KMd)$, where $M$ denotes the number of temporal nodes $V^T$.
The time complexity of each AP block is $\mathcal{O}(Md + Md^2)$, where the first part refers to the aggregation-propagation process and the second part refers to the feature transformation process.
Overall, the time complexity of the TAP-GNN is $\mathcal{O}(MK(d + d^2) + Mld^2))$.
{The complexity comparison between TAP-GNN and TGAT~\cite{tgat_iclr20} is discussed in Appendix~\ref{appendix:tgat}.}

\begin{algorithm}[tbp]
	\caption{Online Inference in Graph-Stream Scenario}
	\label{alg:online}
	\begin{flushleft}
		\textbf{Input}: Stream graph $G'(V',E', t)$; Input features $\{x_v, \forall v \in V'\}$;
		\newline Temporal activation functions $\mathcal{A}_k(\cdot, \cdot)$;
		\newline Aggregation functions $\text{AGG}_k, \forall k \in \{1,\cdots,K\}$;
		\newline Propagation functions $\text{PROP}_k, \forall k \in \{1,\cdots,K\}$;
		\newline Historical node embeddings $\textbf{H}_k(V'), \forall k \in \{1, \cdots, K\}$;
		\newline
		\textbf{Output}: Updated node embeddings $\textbf{H}_k^t(V'), \forall k \in \{1, \cdots, K\}$.
	\end{flushleft}
	\begin{algorithmic}[1]
		\FOR {$k=1 \cdots K$}
			\STATE $\textbf{H}_{k-1}^t(V') \leftarrow \{ \mathcal{A}_k(\textbf{H}_{k-1}(v), t), \forall v \in V' \}$;
			\STATE $\textbf{A}_{k}^t(V') \leftarrow \text{AGG}_k(E', \textbf{H}_{k-1}(V'))$;
		\ENDFOR
		\STATE Construct $L^T = \{ \{v_{t^-}, v_t\}, \forall v \in V'\}$.
		\FOR {$k=1 \cdots K$}
			\STATE $\textbf{A}_{k}(V') \leftarrow \text{CONCAT}(\textbf{H}_k(V'), \textbf{A}_{k}^t(V'))$
			\STATE $ \textbf{H}_{k}^t(V') \leftarrow \text{PROP}_k(L^T, \textbf{A}_{k}(V'))$;
		\ENDFOR
		\STATE return $\textbf{H}_k^t(V'), \forall k \in \{1, \cdots, K\}$.
	\end{algorithmic}
\end{algorithm}

\subsection{Online Inference in Graph-Stream Scenario}

Temporal graphs usually evolve with continuously occurring edges, called graph-stream.
It is essential to adapt the node embedding model for the graph-stream scenario to capture the node dynamics as soon as possible.
Given a batch of edges occurring {simultaneously}, our proposed TAP-GNN can be naturally extended to update the embeddings of the affected nodes efficiently.
As illustrated in Section~\ref{section:ap-block}, the affected nodes only exist in the upcoming batch of edges, and they don't influence the historical neighbors due to the temporal order. 
Then, the affected nodes update their embeddings by fusing the historical embeddings with current embeddings.

Let $V'$ be the affected nodes, $E'$ be the batch edges, $t$ be the time point, {and} current node embeddings can be easily computed using message-passing GNNs as illustrated in Algorithm~\ref{alg:online}. 
Only $V'$ and $E'$ are involved in the computation of current embeddings.
In the graph-stream scenario, each of the propagation link lists contains two elements: the latest node embeddings and current node embeddings, which can be implemented efficiently.
To compute the updated node embeddings, the historical embeddings of each layer have to be saved and fused with the propagation functions.
In general, each batch of new interactions updates the embeddings of the paired nodes accordingly. 

\textbf{Complexity analysis.}
The online TAP-GNN shares the same parameters with the original TAP-GNN, which is also $\mathcal{O}(Kd^2 + ld^2)$.
Let $n$ denote the size of $V'$, and $m$ denote the size of $E'$.
The online TAP-GNN only needs the latest node embeddings so that its space complexity is $\mathcal{O}(KNd)$, where $K$ is the number of AP blocks, $N$ is the size of total nodes, and $d$ is the dimension of embeddings.
The time complexity of computing current embeddings (line 1 to line 4 in Algorithm~\ref{alg:online}) is $\mathcal{O}(Kmd + Knd^2)$, where the first part is the message-passing process and the second part is the feature transformation process.
The time complexity of fusing historical embeddings and current embeddings is $\mathcal{O}(nd)$.

Overall, the time complexity of online inference of TAP-GNN is $\mathcal{O}(K(md + nd^2) + nd)$.

\subsection{{Discussions with TDGNN~\cite{tdgnn} and TGAT~\cite{tgat_iclr20}}}

{

Existing temporal GNNs~\cite{tdgnn,tgat_iclr20} design specific aggregation operations for neighbors, injecting temporal information into node embeddings.
Firstly, their aggregation operations work on a much smaller neighborhood than that of our TAP-GNN, which implies that TDGNN~\cite{tdgnn} and TGAT~\cite{tgat_iclr20} lose neighborhood information before their GNNs.
Secondly, the scalability of TDGNN~\cite{tdgnn} and TGAT~\cite{tgat_iclr20} is restricted since the computation complexity increases exponentially with the number of layers.
In contrast, TAP-GNN could flexibly scale to large-scale temporal graphs due to its linear growth of computation complexity with the number of layers.

}

\begin{table*}[ht]
	\caption{Temporal graph data and statistics. $\vert V \vert$ is the number of nodes in the dataset. $\vert E \vert$ is the number of temporal edges. $d_{\text{max}}$ is the maximum temporal node degree. $\bar{d}$ is the average temporal node degree.}
	\centering
	\begin{tabular}{lccrrrlrrr}
		\toprule
		Temporal Graph       & Bipartite & Edge Type & Repetition  & $\vert V \vert$ & $\vert E \vert$  & Density & $d_{\text{max}}$ & $\bar{d}$ & Timespan (days) \\
		\midrule
		\multicolumn{10}{c}{Future Link Prediction} \\
		\midrule
		fb-forum            & False &  Post      & 20.8\%  & 899             & 33.7K             & 0.08    & 1.8K             & 37.51     & 164.49          \\
		soc-sign-bitcoinotc & False & Trans     & 0.0\%  & 5.8K            & 35.5K           & 0.002   & 1.2K             & 6.05      & 1903.27         \\
		ia-escorts-dynamic & True & Rating    & 10.9\%  & 10K             & 50.6K              & 0.009   & 616              & 5.01      & 2232.00         \\
		ia-movielens-user2tags-10m & True   & Assignment & 19.9\%  & 17K             & 95.5K & 0.0007        & 6.0K             & 5.8       & 1108.97         \\
		soc-wiki-elec     & False  & Election   & 0.2\%        & 7.1K            & 107.0K    & 0.004        & 1.3K             & 15.04     & 1378.84         \\
		ia-slashdot-reply-dir & False  & Post  & 4.2\%      & 51K             & 140.7K          & 0.0001  & 3.3K             & 2.76      & 977.36          \\
		\midrule
		\multicolumn{10}{c}{Dynamic Node Classification} \\
		\midrule
		Wikipedia	& True & Edit & 79.1\%  & 9.2K          & 157.4K      & 0.0036  & 1.9K          &  57.64     &  29.77        \\
		Reddit		& True & Post & 61.4\%	&  10.9K         &    672.4K      & 0.011  & 58.7K  &       61.22      &   31.00    \\
		\bottomrule
	\end{tabular}
	\label{tab:data}
\end{table*}

\section{Experiment}

We conduct the experiments over a wide range of temporal dynamic graphs to investigate the effectiveness of our proposed \textbf{TAP-GNN} framework.
The statistics summarization of temporal graphs is listed in Table~\ref{tab:data}.
The configurations of the competing baselines and our method are described in detail.
Two experiment tasks, future link prediction and dynamic node classification, are designed to demonstrate the effectiveness of our proposed TAP-GNN framework.
Discussions on the experimental results and ablation studies are provided in the following.

\subsection{Datasets and Evaluation}
\subsubsection{Datasets}

The datasets listed in Table~\ref{tab:data} are used for two evaluation tasks. 
Datasets for future link prediction are collected from Network Repository~\cite{network-data}\footnote{http://networkrepository.com/dynamic.php}.
Datasets for dynamic node classification are obtained from SNAP~\cite{kumar2019jodie}\footnote{http://snap.stanford.edu/jodie/}.
They are divided into two categories because of their different repetition ratios, which means that a node interacted with the same node consecutively.
Temporal graphs with high repetition ratios usually follow regular behavior patterns and are not suitable for the future link prediction task.
In addition, datasets for future link prediction also vary greatly in their graph properties, which is challenging for model robustness.

\textbf{Future link prediction}. 
These temporal graphs are provided without node or edge features from various fields, from Facebook posts (e.g., fb-forum) to Wikipedia {elections} (e.g. soc-wiki-elec). 
They differ significantly in many fields: the smallest node size is 899, and the largest one is over 51 thousand; two of the datasets are bipartite, and the others are homogeneous; the timespan of them starts from 164 days to 2232 days when setting the time unit as one day, etc.

\begin{itemize}
	\item \textbf{fb-forum:} The network shows user activities on the Facebook-like forum network {with} 899 user nodes and 33 thousand posts. 
	\item \textbf{soc-sign-bitcoinotc:} This is a signed trust network on a platform called Bitcoin OTC\footnote{https://www.bitcoin-otc.com/}, where people trade using Bitcoin. It is noteworthy that no repeated interactions happen in this network. The network contains 5.8 thousand users and 35.5 thousand trust ratings.

	\item \textbf{ia-escorts-dynamic:} The network is a bipartite rating network, containing more than 10 thousand nodes and 50 thousand ratings. 
	\item \textbf{ia-movielens-user2tags-10m:} This is another bipartite network from the collected MovieLens dataset\footnote{https://movielens.org/}, where edges connect users with the tags they gave to the movies. The sparse network has over 17 thousand nodes and 95 thousand edges.
	\item \textbf{soc-wiki-elec:} The network presents the Wikipedia administrator election and voting history, which contains more than 7 thousand users and 107 thousand interactions. 
	\item \textbf{ia-slashdot-reply-dir:} This is a reply network collected from the technology website Slashdot over 977 days. The network has the largest nodes and edges for future link prediction, consisting of 51 thousand users and over 140 thousand replies.
\end{itemize}

\textbf{Dynamic node classification}. The additional two datasets are used for dynamic node classification with the extremely imbalanced class ratio. The features of user edits in Wikipedia and user posts {on} Reddit are both converted into the 172-dimensional vectors under the \textit{linguistic inquiry and word count} (LIWC) categories~\cite{liwc2001}.
\begin{itemize}

	\item \textbf{Wikipedia:} 
	The network consists of 8,227 users and 1,000 most edited pages, with 157,474 interactions in total. 
	The timestamp of interaction tells when the user edits the page. 
	There are 217 positive labels (0.41\%) among all interactions, which tells whether the user is banned from editing the Wikipedia page.
	\item \textbf{Reddit:} 
	The network contains 10,000 active users and 1,000 active subreddits, resulting in 672,447 temporal edges. 
	Each edge with a timestamp means a user made a Reddit post at some time. 
	There are 366 positive labels (0.05\%) among all posts, which indicates the user is banned from posting under the subreddit.
\end{itemize}

Overall, the variety of datasets could fully demonstrate the difference and robustness of compared methods.

\subsubsection{Evaluation Tasks}

\textbf{Evaluation metrics.} 
We adopt the Accuracy metric and the area under the receiver operating characteristic curve (AUC-ROC) as the evaluation metrics for different tasks. 

\textbf{Future link prediction}. For methods comparison, we perform future link prediction on a hold-out evaluation dataset and run each experiment five times to ensure performance robustness. We select the hyperparameters of the best average performance on the validation dataset for each method and report the corresponding performance on the test dataset. The dataset is sorted by the timestamp ascendingly, and the first 70\% of edges are taken for training, and the next 15\% are used for validation. Since the datasets don't have {a} node or edge features, we remove edges with unseen nodes in the test dataset for correctness. The left edges in the test dataset are used as positive samples. We sample an equal number of edges as negative samples by substituting the target node of the positive edges with a randomly sampled node.
For each dataset, we generate a labeled test set for future link prediction. We use the training set to generate embeddings for each node for models that are not directly designed for future link prediction. Then for each prediction edge, the embeddings of two nodes are concatenated as its features. We apply the same strategy to construct a labeled training set and use logistic regression (LR) to perform future link prediction. For end-to-end models, we measure the reported performance of their output probabilities. The evaluation metric is AUC-ROC and Accuracy.

\textbf{Dynamic node classification}. 
Besides the future link prediction task, we introduce the dynamic node classification task as described in~\cite{tgat_iclr20}. 
The task is challenging due to the extremely imbalanced labels: \textit{Wikipedia} has 217 positive labels with 157,474 interactions (=0.14\%), while \textit{Reddit} has 366 true labels among 672,447 interactions (=0.05\%). 

The temporal graphs are split chronologically into 70\%-15\%-15\% for training, validation, and testing. 
Since removing unseen nodes from the test set will largely reduce the positive samples, we only compare against methods that can generalize to unseen nodes.
We train the self-supervised future link prediction task firstly and use an MLP classifier to predict the dynamic node label with the temporal node embeddings. 
Notably, we adopt the same concatenation function on the two interaction node embeddings for prediction as TGAT~\cite{tgat_iclr20}. 
The evaluation metric is only AUC-ROC due to the highly imbalanced ratio.

\subsection{Experiment Setting}

\begin{table*}[htbp]
	\caption{{Experimental results of future link prediction.}}
	\centering
	\resizebox{\textwidth}{!}{
	\begin{tabular}{l|c|c|c|c|c|c}
		\toprule
		\textbf{Temporal Graph}   & \multicolumn{2}{c|}{\textbf{fb-forum}} & \multicolumn{2}{c|}{\textbf{soc-sign-bitcoinotc}} & \multicolumn{2}{c}{\textbf{ia-escorts-dynamic}} \\  
		\midrule
		\textbf{Metrics} & AUC & Accuracy & AUC & Accuracy & AUC & Accuracy \\
		\midrule
		\textbf{Node2Vec}\cite{grover2016node2vec} & $ 0.823 \pm 0.008 $ & $ 0.744 \pm 0.009 $ & $ 0.774 \pm 0.012 $ & $ 0.708 \pm 0.014 $ & $ 0.839 \pm 0.005 $ & $ 0.769 \pm 0.004 $\\
		\textbf{GraphSAGE}~\cite{hamilton2017graphsage} & $ 0.724 \pm 0.010 $ & $ 0.636 \pm 0.006 $ & $ 0.734 \pm 0.009 $ & $ 0.651 \pm 0.008 $ & $ 0.847 \pm 0.005 $ & $ 0.691 \pm 0.040 $\\
		$\textbf{GraphSAGE}^\star$~\cite{hamilton2017graphsage} & $ 0.826 \pm 0.002 $ & $ 0.750 \pm 0.002 $ & $ 0.858 \pm 0.002 $ & $ 0.803 \pm 0.001 $ & $ 0.888 \pm 0.001 $ & $ 0.824 \pm 0.002 $\\
		\textbf{CTDNE}\cite{nguyen2018ctdne}& $ 0.817 \pm 0.005 $ & $ 0.745 \pm 0.007 $ & $ 0.836 \pm 0.003 $ & $ 0.778 \pm 0.004 $ & $ 0.898 \pm 0.001 $ & $ 0.825 \pm 0.004 $\\
		\textbf{HTNE}\cite{zuo2018htne} & $ 0.715 \pm 0.005 $ & $ 0.668 \pm 0.004 $ & $ 0.639 \pm 0.005 $ & $ 0.611 \pm 0.009 $ & $ 0.810 \pm 0.005 $ & $ 0.749 \pm 0.007 $\\
		\textbf{TNODE}\cite{tnode} & $ 0.795 \pm 0.007 $ & $ 0.716 \pm 0.004 $ & $ 0.793 \pm 0.008 $ & $ 0.724 \pm 0.007 $ & $ 0.872 \pm 0.004 $ & $ 0.801 \pm 0.005 $\\
		\textbf{TGAT}\cite{tgat_iclr20} & $ 0.878 \pm 0.001 $ & $ 0.794 \pm 0.001 $ & $ 0.896 \pm 0.001 $ & $ 0.823 \pm 0.001 $ & $ 0.941 \pm 0.001 $ & $ 0.874 \pm 0.001 $\\
		{\textbf{TDGNN}}\cite{tdgnn} & $ {0.744 \pm 0.014} $ & $ {0.689 \pm 0.052} $ & $ {0.648 \pm 0.038} $ & $ {0.314 \pm 0.246} $ & $ {0.765 \pm 0.013} $ & $ {0.746 \pm 0.009} $\\
		{\textbf{JODIE}}\cite{kumar2019jodie} & $ {0.815 \pm 0.000} $ & $ {0.767 \pm 0.000} $ & $ {0.912 \pm 0.000} $ & $ {0.848 \pm 0.000} $ & $ {0.942 \pm 0.000} $ & $ {0.855 \pm 0.000} $\\
		{\textbf{TGN}}\cite{tgn2020rossi} & $ {0.889 \pm 0.016} $ & $ {0.783 \pm 0.025} $ & $ {0.942 \pm 0.004} $ & $ {0.863 \pm 0.004} $ & $ {0.941 \pm 0.001} $ & $ {0.870 \pm 0.006} $\\
		\textbf{TAP-GNN} & $ \bm{ 0.909 \pm 0.002 } $ & $ \bm{ 0.828 \pm 0.008 } $ & $ \bm{ 0.960 \pm 0.001 } $ & $ \bm{ 0.872 \pm 0.007 } $ & $ \bm{ 0.964 \pm 0.001 } $ & $ \bm{ 0.879 \pm 0.008 } $\\
		\hline
		\textbf{{TAP-GNN*}} & $ {0.889 \pm 0.023} $ & $ {0.818 \pm 0.033} $ & $ {0.965 \pm 0.016} $ & $ {0.857 \pm 0.069} $ & $ {0.966 \pm 0.012} $ & $ {0.900 \pm 0.043} $\\

		\toprule
		\textbf{Temporal Graph}  & \multicolumn{2}{c|}{\textbf{ia-movielens-user2tags-10m}} & \multicolumn{2}{c}{\textbf{soc-wiki-elec}} & \multicolumn{2}{c}{\textbf{ia-slashdot-reply-dir}}\\  
		\midrule
		\textbf{Metrics} & AUC & Accuracy & AUC & Accuracy & AUC & Accuracy \\
		\midrule
		\textbf{Node2Vec}\cite{grover2016node2vec} & $ 0.756 \pm 0.003 $ & $ 0.696 \pm 0.003 $ & $ 0.508 \pm 0.017 $ & $ 0.512 \pm 0.015 $ & $ 0.800 \pm 0.003 $ & $ 0.723 \pm 0.003 $\\ 
		\textbf{GraphSAGE}~\cite{hamilton2017graphsage} & $ 0.800 \pm 0.004 $ & $ 0.710 \pm 0.012 $ & $ 0.577 \pm 0.023 $ & $ 0.563 \pm 0.014 $ & $ 0.782 \pm 0.005 $ & $ 0.644 \pm 0.017 $\\
		$\textbf{GraphSAGE}^\star$~\cite{hamilton2017graphsage} & $ 0.864 \pm 0.002 $ & $ 0.793 \pm 0.001 $ & $ 0.889 \pm 0.010 $ & $ 0.831 \pm 0.013 $ & $ 0.571 \pm 0.038 $ & $ 0.497 \pm 0.012 $\\
		\textbf{CTDNE}\cite{nguyen2018ctdne} & $ 0.796 \pm 0.006 $ & $ 0.728 \pm 0.004 $ & $ 0.606 \pm 0.012 $ & $ 0.558 \pm 0.009 $ & $ 0.851 \pm 0.002 $ & $ 0.782 \pm 0.002 $\\ 
		\textbf{HTNE}\cite{zuo2018htne} & $ 0.736 \pm 0.003 $ & $ 0.694 \pm 0.005 $ & $ 0.535 \pm 0.009 $ & $ 0.536 \pm 0.007 $ & $ 0.691 \pm 0.002 $ & $ 0.660 \pm 0.003 $\\ 
		\textbf{TNODE}\cite{tnode}  & $ 0.793 \pm 0.013 $ & $ 0.713 \pm 0.012 $ & $ 0.613 \pm 0.013 $ & $ 0.566 \pm 0.007 $ & $ 0.804 \pm 0.008 $ & $ 0.731 \pm 0.016 $\\
		\textbf{TGAT}\cite{tgat_iclr20} & $ 0.884 \pm 0.001 $ & $ 0.820 \pm 0.001 $ & $ 0.869 \pm 0.003 $ & $ 0.820 \pm 0.005 $ & $ 0.805 \pm 0.001 $ & $ 0.676 \pm 0.001 $\\
		{\textbf{TDGNN}}\cite{tdgnn} & $ {0.670 \pm 0.015} $ & $ {0.648 \pm 0.023} $ & $ {0.503 \pm 0.038} $ & $ {0.370 \pm 0.084} $ & $ {0.777 \pm 0.023} $ & $ {0.664 \pm 0.205} $\\
		{\textbf{JODIE}}\cite{kumar2019jodie} & $ {0.862 \pm 0.006} $ & $ {0.775 \pm 0.004} $ & $ {0.912 \pm 0.000} $ & $ {0.848 \pm 0.000} $ & $ {0.942 \pm 0.000} $ & $ {0.855 \pm 0.000} $\\
		{\textbf{TGN}}\cite{tgn2020rossi} & $ \bm{0.921 \pm 0.004} $ & $ \bm{0.841 \pm 0.004} $ & $ {0.769 \pm 0.028} $ & $ {0.705 \pm 0.014} $ & $ {0.896 \pm 0.005} $ & $ {0.807 \pm 0.025} $\\	
		\textbf{TAP-GNN} & $ {{0.913 \pm 0.001}} $ & $ {{0.835 \pm 0.004}} $ & $ \bm{ 0.989 \pm 0.001 } $ & $ \bm{ 0.978 \pm 0.002 } $ & $ \bm{ 0.980 \pm 0.000 } $ & $ \bm{ 0.944 \pm 0.001 } $\\
		\hline
		\textbf{{TAP-GNN*}} & $ {0.923 \pm 0.032} $ & $ {0.836 \pm 0.091} $ & $ {0.989 \pm 0.003} $ & $ {0.978 \pm 0.004} $ & $ {0.982 \pm 0.007} $ & $ {0.954 \pm 0.020} $\\
		\bottomrule
	\end{tabular}
	}
	\label{tab:comp}
\end{table*}

\subsubsection{Baseline Methods}

We evaluate the TAP-GNN framework for learning temporal node embeddings against the following baseline methods. We set the embedding dimension $D$ as $128$ for all compared methods for a fair comparison. Additionally, all methods use the one-hot encoding features and generate temporal embeddings according to the method settings.

\begin{itemize}
	\item Node2Vec~\cite{grover2016node2vec} extends DeepWalk~\cite{perozzi2014deepwalk} with two hyper-parameters to explore the neighborhood more flexibly, which is a strong baseline in skip-gram based models. However, random walks defined on static graphs cannot be generalized to temporal graphs directly. 
	\item GraphSAGE~\cite{hamilton2017graphsage} adopts several aggregators for neighbor messages and introduces an inductive learning paradigm for large-scale graphs, which is a strong baseline in static GNNs.
	\item $\text{GraphSAGE}^\star$~\cite{hamilton2017graphsage} is an extended version of GraphSAGE with the temporal subgraph batching strategy described in TGAT~\cite{tgat_iclr20}. The uniform sampling strategy is applied here.
	\item CTDNE~\cite{nguyen2018ctdne} generates the final state for each node in temporal graphs with the temporal random walks. However, it is challenging to deploy CTDNE for an online inference scenario. 
	\item HTNE~\cite{zuo2018htne} fuses the historical neighbor information with the Hawkes process and an attention mechanism, which is implemented in a recurrent paradigm. Despite its efficiency, the neighbor window of HTNE limits its receptive field.
	\item TNODE~\cite{tnode} aligns the consecutive embeddings with a rotation matrix and uses an LSTM to generate evolving embeddings based on Node2Vec~\cite{grover2016node2vec}. The algorithm is efficient but cannot generalize for online inference. 
	\item TGAT~\cite{tgat_iclr20} incorporates a time encoding kernel and an attention mechanism into dynamic node embeddings based on the sampled temporal subgraph. TGAT may fail in some cases due to the lack of the whole neighborhood, as shown in our experiments.
	\item {JODIE~\cite{kumar2019jodie} is built upon recurrent networks and TDGNN~\cite{tdgnn} designs various kernels of GNNs. Furthermore, TGN~\cite{tgn2020rossi} adopts advantages both of recurrent networks and GNNs.}
\end{itemize}

\subsubsection{Experiment Setup}

\textbf{Static Methods Configuration.} Static graph methods use the training data both for training and test. For \textit{Node2Vec}, we grid search over $p, q \in \{ 0.25, 0.50, 1, 2, 4 \}$ and set the other hyperparameters as default in~\cite{grover2016node2vec}. For \textit{GraphSAGE}, a trainable node embedding matrix is employed instead of node features due to the lack of node features. The training adjacency matrix is used in the test phase to avoid information leakage. The other hyper-parameters of \textit{GraphSAGE} are set as default in~\cite{hamilton2017graphsage}.

\textbf{Temporal Methods Configuration.} For temporal graph methods based on graph snapshots, we run methods over different graph snapshot divisions, namely $\{8,32,128\}$ snapshots, and report the best test performance over these three kinds of division strategies. For \textit{CTDNE}, we use the learned node embeddings during training for the test. For \textit{HTNE}, we grid search the history length over $\{10, 20, 30\}$ as suggested in~\cite{zuo2018htne} and report the test performance. \textit{tNodeEmbed}~\cite{tnode} is an end-to-end framework that utilizes RNN over embeddings for various snapshots so that we run this method over three different division strategies. TGAT~\cite{tgat_iclr20} is capable of generating dynamic node representations and {is} well suitable for future link prediction. The input features are one-hot encodings instead of all-zero vectors, and the sampling strategies are evaluated under uniform sampling and inverse temporal sampling. Other hyper-parameters of TGAT are set as default in~\cite{tgat_iclr20}. 
{Moreover, we search hyper-parameters of JODIE~\cite{kumar2019jodie}, TDGNN~\cite{tdgnn}, and TGN~\cite{tgn2020rossi} according to their papers.}

\textbf{TAP-GNN Configuration.} The input features are one-hot encoding features, and the hidden dimension is all set as 128 for a fair comparison. If edge features are given, node features, edge features, and time encodings are concatenated for the next layer, including {the} AP block and the projection layer. The whole framework is described as shown in Fig.~\ref{fig:framework}, where AP blocks are stacked two times, and the projection layer is a two-layer MLP.   
{Moreover, we implement TAP-GNN* in Table~\ref{tab:comp} by aggregating directly from the whole neighborhood and reusing the trained weights of TAP-GNN to validate whether the intermediate propagation {affects} the prediction performance.}

\textbf{Implementation.} Our proposed framework is implemented using PyTorch~\cite{pytorch2019paszke} and DGL (Deep Graph Library)~\cite{dgl2019wang}. One-hot encodings are set for temporal graphs without node features, where the embedding matrix is initialized with Glorot initialization. We adopt the Adam optimizer, and select the learning rate as 0.01 from $\{0.01, 0.001, 0.0001\}$. The default batch size is set {at} 256, the default dropout ratio is set {at} 0.2, and the negative samples are set {at} 1 for each positive sample. The early-stopping strategy is applied during training until the validation \textsl{AUC} score does not improve over five epochs. Our experimental machine is a 24-core Linux server with 125GB memory and a Quadro P6000 GPU.

\textbf{Downstream node classification.} 
We use the same three-layer MLP~\cite{tgat_iclr20} as a classifier. 
Since the banned user is reported with the corresponding wiki page/subreddit, the two-node embeddings are concatenated as input features for dynamic node classification. 
{The} MLP classifier is trained with the Adam optimizer and the Glorot initialization, and the early-stopping strategy with ten epochs. 
Due to the data imbalance, the positive labels are oversampled to achieve the balance label ratio in each batch.

\begin{table}[t]
    \caption{AUC scores for dynamic node classification. The \textbf{Bold} font and the \underline{underline} font refer to the best and second-best performance.}
    \centering
    \footnotesize{
    \resizebox{.4\textwidth}{!}{
    
    \begin{tabular}{rll}
        \toprule
        Methods & Wikipedia & Reddit \\
        \midrule
        $\textbf{GraphSAGE}^\star$\cite{hamilton2017graphsage} & $82.4 \pm 0.7$ & $61.2 \pm 0.6$ \\
        \textbf{CTDNE}\cite{nguyen2018ctdne} & $75.9 \pm 0.5$ & $59.4 \pm 0.6$ \\
        \textbf{TGAT}\cite{tgat_iclr20} & $83.7 \pm 0.7$ & $65.6 \pm 0.7$ \\
		{\textbf{JODIE}}\cite{kumar2019jodie} & ${83.2 \pm 0.5}$ & ${59.9 \pm 2.1}$ \\
		{\textbf{TGN}}\cite{tgn2020rossi} & ${\bm{87.8 \pm 0.2}}$ & ${\bm{67.1 \pm 0.9}}$ \\
        \textbf{TAP-GNN} & $\underline{85.5 \pm 0.3}$ & $\underline{66.3 \pm 0.5}$ \\
        \bottomrule
    \end{tabular}}
    }
    \label{tab:nc}
\end{table}

\subsection{Effectiveness of TAP-GNN}

\subsubsection{Future Link Prediction}

Table~\ref{tab:comp} presents the results of the future link prediction of our TAP-GNN, two static graph methods, and five temporal graph methods. 
The proposed TAP-GNN outperforms other state-of-the-art baselines on all experimental temporal graphs, which achieves at most 12.9\% improvement of AUC and 16.2\% improvement of Accuracy on \emph{ia-slashdot-reply-dir}. 
These datasets can be divided into two categories according to their repetition ratios and experimental performance of compared methods: high-repetition datasets (i.e., fb-forum, ia-escorts-dynamic and ia-movielens-user2tags-10m) and low-repetition datasets (i.e., soc-sign-bitcoinotc, soc-wiki-elec and ia-slashdot-reply-dir).
TAP-GNN shows much better improvements on low-repetition datasets than on high-repetition datasets because of its ability to handle the whole neighborhood.
Meanwhile, most compared methods can hardly handle the varying neighborhood and graph stream simultaneously.
Overall, the TAP-GNN framework demonstrates both superior and robust performance. 
It indicates that the varying neighborhood plays an essential role in the task of predicting future link existence.
{
TAP-GNN* in Table~\ref{tab:comp} performs on par with TAP-GNN, implying that differences between direct aggregation and TAP-GNN are probably caused by calculation errors.
}

For static graph methods, Node2Vec performs better on three datasets, while GraphSAGE beats Node2Vec on the left three datasets.
It implies that it is essential to balance the structure information and local neighbor information since Node2Vec embeds the former, and GraphSAGE encodes the latter.
It is interesting to see that $\text{GraphSAGE}^\star$ outperforms GraphSAGE by a large margin on five datasets but fail greatly on \emph{ia-slashdot-reply-dir}.
The former tells that node activity is positively related to their recent interactions on most temporal graphs, designed by the temporal negative sampling loss.
However, the assumption may fail in cases {where} long-range dependencies exist among nodes.

{
For temporal graph methods, TGAT and TGN consistently outperform other baseline methods on most datasets. Nevertheless, these two methods both degrades on two sparse networks \emph{soc-wiki-elec} and \emph{ia-slashdot-reply-dir}. JODIE~\cite{kumar2019jodie} presents much better performance than TGAT and TGN on two sparse networks, indicating the importance of neighborhood for future link prediction. Meanwhile, temporal graph methods including CTDNE, HTNE, TNODE, and TDGNN obtain comparable performance to GraphSAGE*, implying the superiority of GNNs.
}

\subsubsection{Dynamic Node Classification}

Table~\ref{tab:nc} presents the AUC-ROC scores of our TAP-GNN and {five} compared temporal graph methods.
Since the node labels in Wikipedia and Reddit are extremely imbalanced, only methods that can generalize to unseen nodes are shown here.
Also, the Accuracy metric is omitted here because of the imbalanced label ratio.
The positive labels in both datasets denote that users are banned from the destination wiki pages/subreddits.
Intuitively, this task reflects the flexibility of experimental models because users exhibit frequent interactions with low banned times.
{
Table~\ref{tab:nc} shows that both TGN and TAP-GNN outperform other baselines, and TGN performs better than TAP-GNN. We conjecture that the frequent update of node memory contributes to the superiority of TGN, which motivates us to design a more flexible graph kernel in future work.  
}

\begin{table}[t]
	\caption{{Time per batch (ms) on Wikipedia against TGAT.} }

    \centering

    \resizebox{0.9\linewidth}{!}{
    
    \begin{tabular}{r|ccccc|ccc}
        \toprule
		\multirow{2}{*}{Methods}& \multicolumn{5}{c|}{Batch-size } & \multicolumn{3}{c}{Layers} \\

		& 100 & 200 & 300 & 400 & 500 & 1 & 2 & 3 \\
        \midrule
        \textbf{TGAT}\cite{tgat_iclr20} &32.6 & 49.7 & 68.7 & 110.2 & 113.9 & 5.6 & 49.7 & 934.7  \\
		\textbf{TAP-GNN} & {11.0} & {12.5} & {13.8} & {15.1} & {16.3} & {5.6} & {12.5} & {13.8}  \\ 
		\textbf{Speedup} &$ {3 \times }$ &  $ {4 \times} $ &  $ {5 \times} $ &  $ {7 \times} $ & ${7 \times} $ & $ {\sim 1}  $ &  $ {4 \times} $ &  $ {68 \times} $  \\
        \bottomrule
	\end{tabular}
	}

    \label{tab:time}
\end{table}
\begin{figure}[t]
    \centering
    \includegraphics[width=.4\textwidth, height=0.2\textwidth]{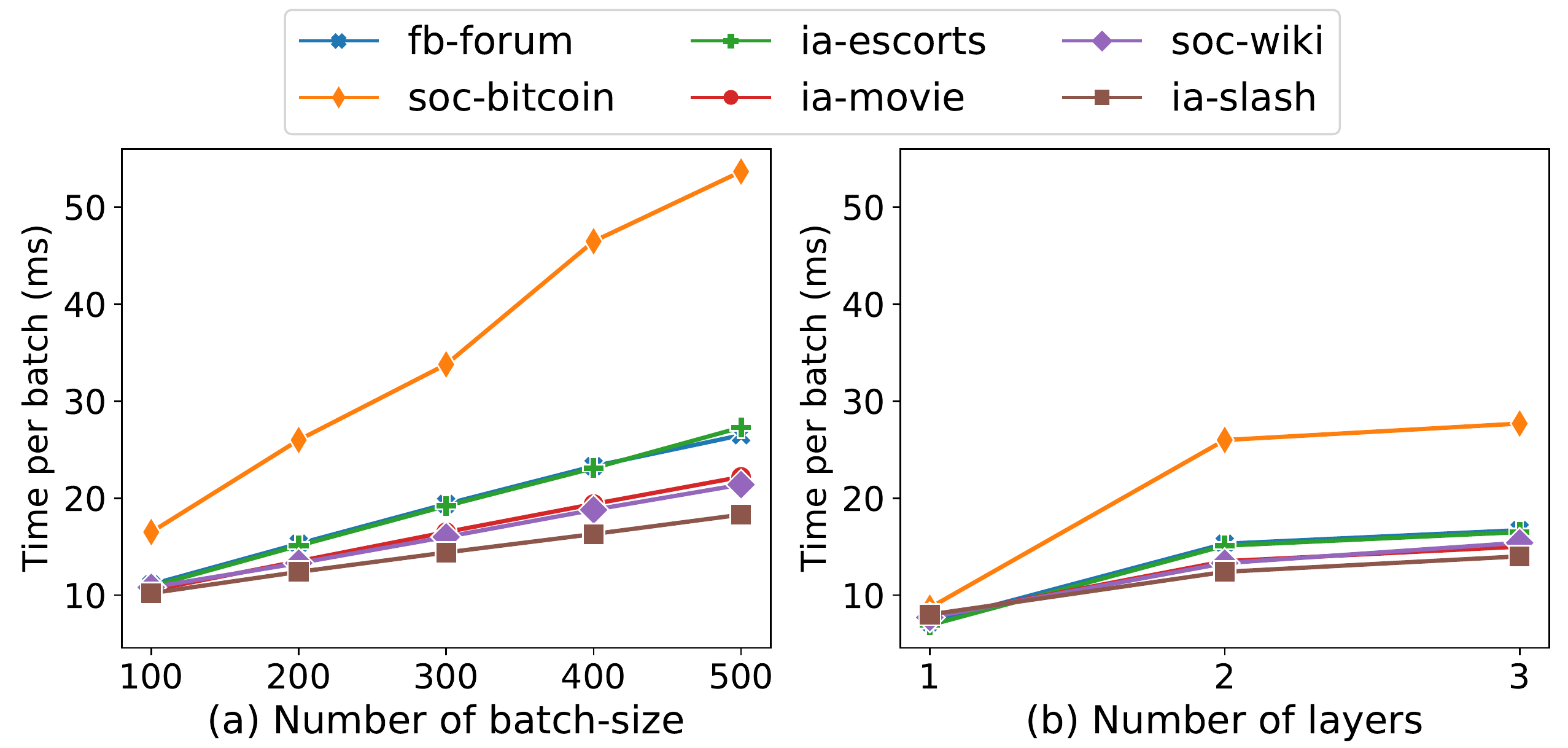}
    \caption{{Time per batch (ms) of TAP-GNN.}}
    \label{fig:time}
\end{figure}

\subsection{Efficiency of TAP-GNN}

The proposed TAP-GNN consists of many specific modules: AP blocks, the temporal activation function, and the projection layer, each with its own hyper-parameters.
Experiments in this section aim to validate the efficiency of different components of TAP-GNN by varying the hyper-parameters.
Without specific settings, the batch size is 256, the number of AP blocks is 2, the embedding dimensionality is set {to} 128, and the dropout ratio is 0.2. 
Also, the names of the datasets are abbreviated without confusion for simplicity in the figures.

\subsubsection{Online Inference Efficiency}

Since other approaches such as CTDNE~\cite{nguyen2018ctdne}, HTNE~\cite{zuo2018htne}, and TNODE~\cite{tnode} are not designed for graph representation in the graph stream scenario, we only compare against TGAT~\cite{tgat_iclr20} to show the online inference efficiency.
{

For a fair comparison, we implement the attention kernel with TAP-GNN since TGAT costs much time in the attention kernel, as shown in Table~\ref{tab:time} and Fig.~\ref{fig:time}.
}
The task predicts the probability {that} the edge exists between two nodes when given a future time point.
The graph topology keeps evolving with a batch of new interactions.
Specifically, the original implementation of TGAT sampling is extremely slow and can achieve {the speedup} nearly 50 times with the Numba\footnote{http://numba.pydata.org/} library, which dramatically reduces the time cost of graph convolutions of TGAT. 
As reported in Table~\ref{tab:time}, the consumed time of TAP-GNN remains nearly constant while TGAT presents the linearly increasing time cost when varying the batch size of the upcoming new interactions.
When changing the number of layers, TGAT demonstrates the exponential growth of inference time, which inherits the disadvantages of GraphSAGE~\cite{hamilton2017graphsage}.
While our proposed TAP-GNN demonstrates the linear growing rate as shown both in Table~\ref{tab:time} and Fig.~\ref{fig:time}.
It is also crucial that the online inference time of TAP-GNN is almost irrelevant to the graph size, which indicates the scalability of TAP-GNN on large industrial temporal graphs.

\subsubsection{Ablation Studies}

\begin{figure}[t]
    \centering
    \includegraphics[width=.4\textwidth, height=0.25\textwidth]{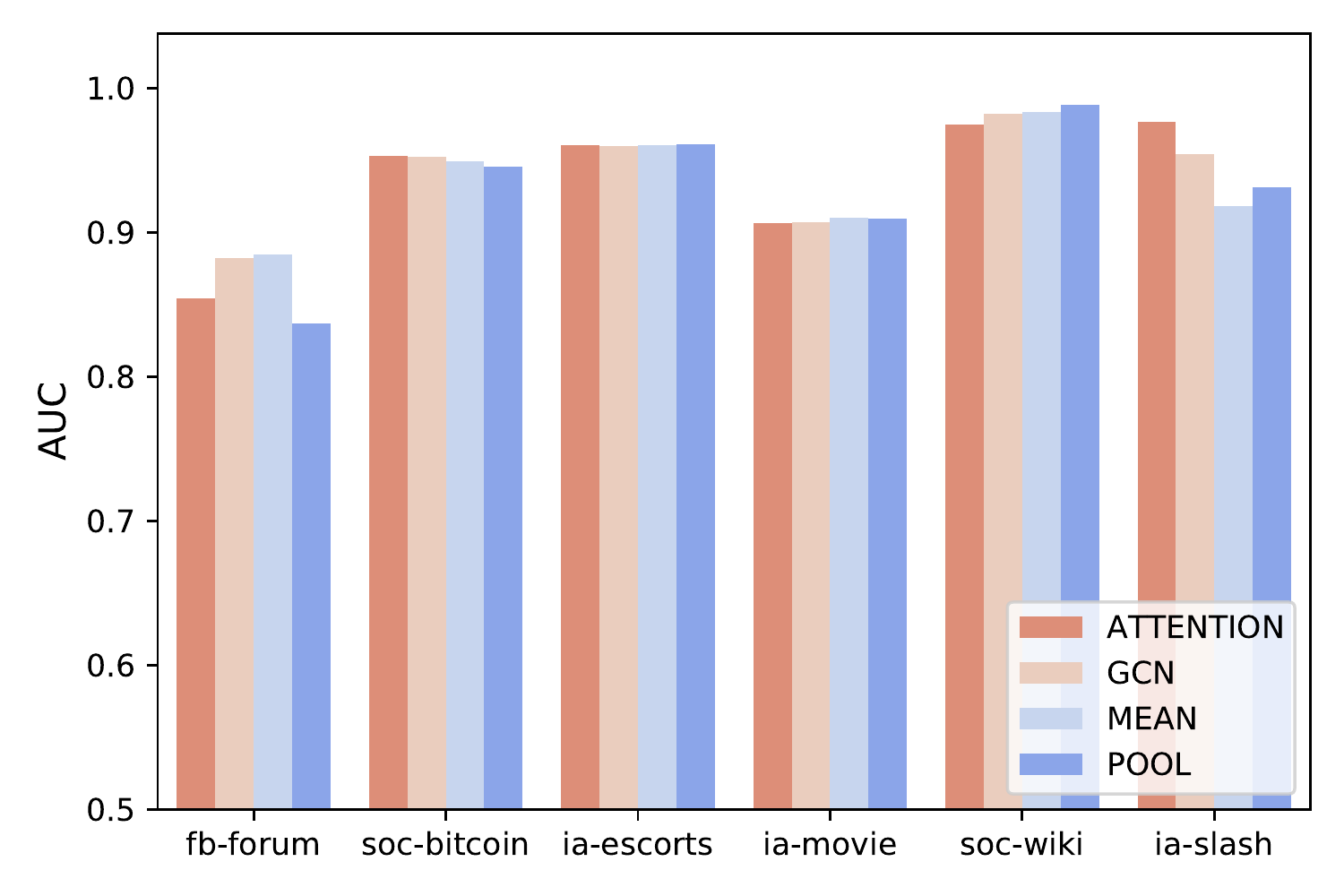}
	\caption{AUC scores under different graph convolution kernels.}
    \label{fig:kernel}
\end{figure}

\begin{figure}[t]
    \centering
    \includegraphics[width=.4\textwidth, height=0.25\textwidth]{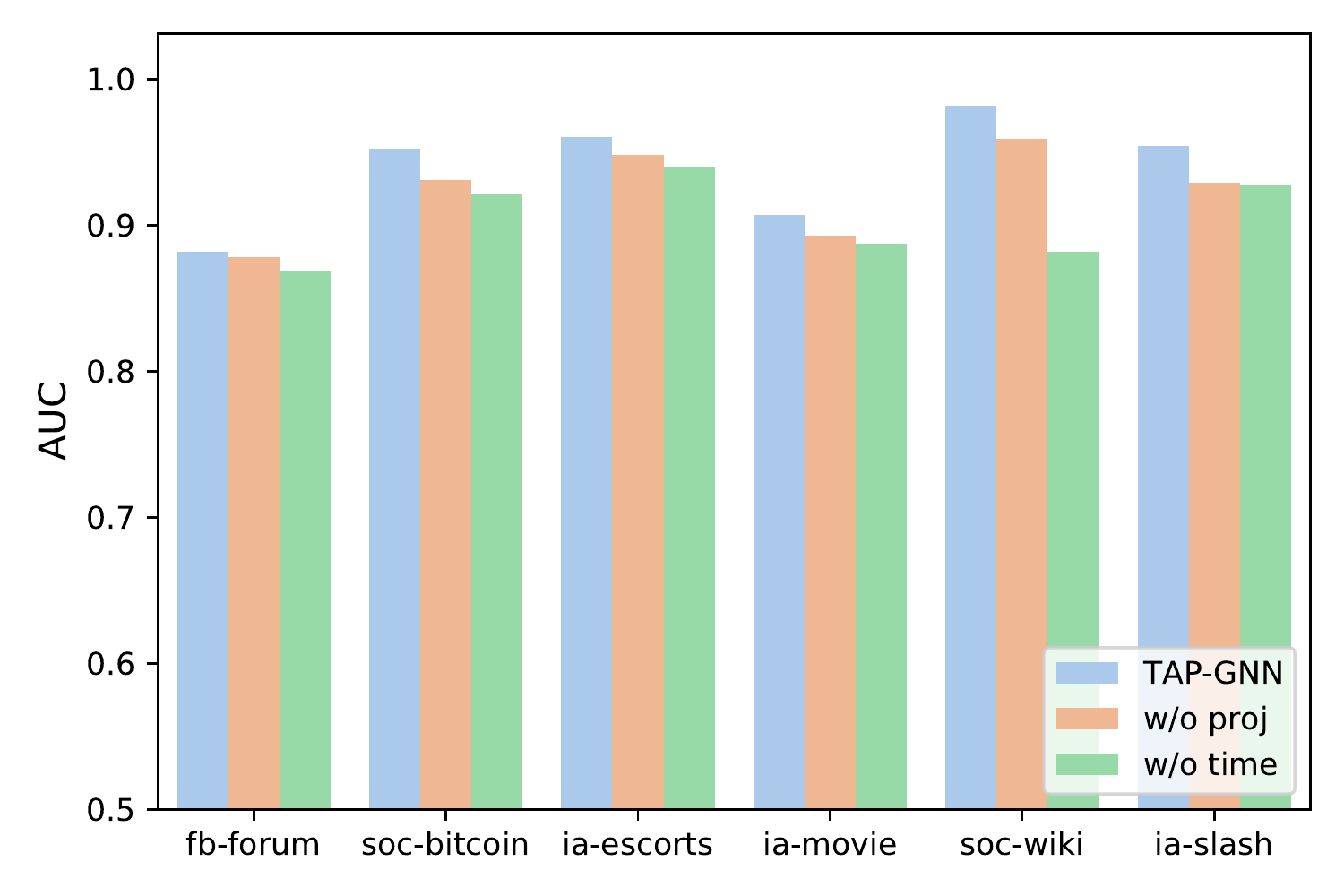}
    \caption{Ablation studies of the projection layer and temporal activation function, where w/o proj denotes TAP-GNN without the projection layer and w/o time denotes TAP-GNN without temporal activation function.}
    \label{fig:ablation}
\end{figure}

\begin{figure}[t]
    \centering
    \includegraphics[width=.45\textwidth, height=0.2\textwidth]{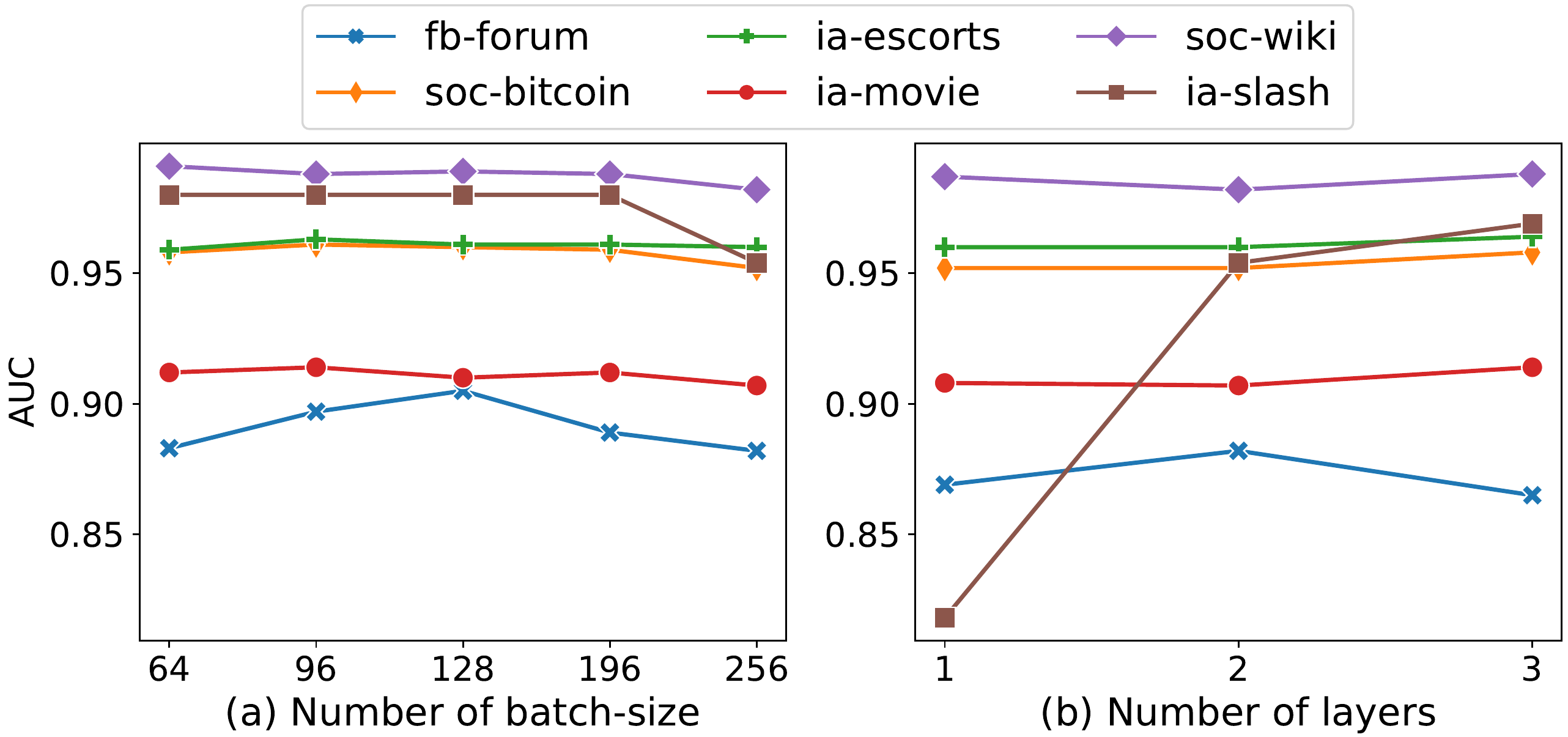}
    \caption{Effects of different batch-sizes and number of layers.}
    \label{fig:batch_layer}
\end{figure}

\begin{figure}[t]
    \centering
    \includegraphics[width=.45\textwidth, height=0.2\textwidth]{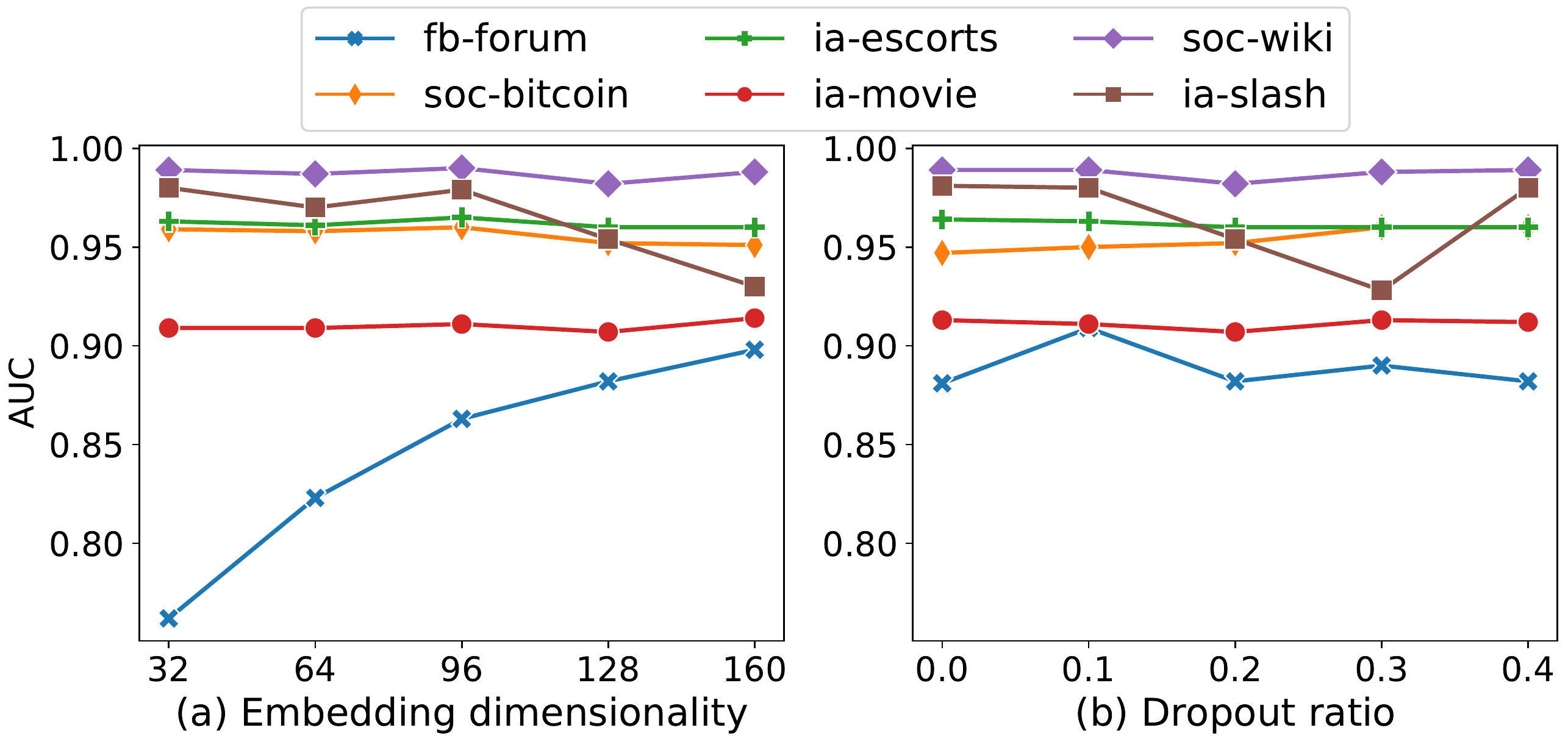}
    \caption{Effects of different embedding dimensionalities and dropout ratios.}
    \label{fig:hidden_dropout}
\end{figure}

\textbf{Graph convolution kernels.} 
{
We implement four graph convolution kernels ATTENTION, GCN, MEAN, and POOL according to Eq.~\ref{eq:graph-kernel} and Appendix~\ref{appendix:attention}.
}

As depicted in Fig.~\ref{fig:kernel}, GCN performs the most robust among all datasets, while {ATTENTION}, MEAN and POOL exhibit performance degradation over several datasets. 
Further, MEAN shows inferior to GCN on large and sparse graphs, and {ATTENTION} and POOL perform worse than GCN on small and dense graphs. 
Hence, we select the GCN as the default graph convolution kernels for the following experiments.
Overall, these three kernels demonstrate comparable performance on most datasets, which implies the robustness of TAP-GNN.

\textbf{Temporal activation function.}
The temporal activation function encodes the time point of each interaction to generate time-respecting node embeddings.
It is useful to capture the periodic information inside the dynamic interactions.
As described in Fig.~\ref{fig:ablation}, TAP-GNN, which is equipped with the temporal activation function, shows significant improvements on the baseline without the function.

\textbf{The projection layer.}
The obtained node embeddings are further projected to the future time point with the projection layer.
TAP-GNN without the projection layer predicts the probability of the edge's existence by a fully-connected layer.
As plotted in Fig.~\ref{fig:ablation}, it is crucial to perform an additional projection on the learned embeddings.

\subsubsection{Parameter Sensitivity}

\begin{figure*}[t]
	 \centering
     \subfloat[][soc-bitcoin]{\includegraphics[width=.25\textwidth]{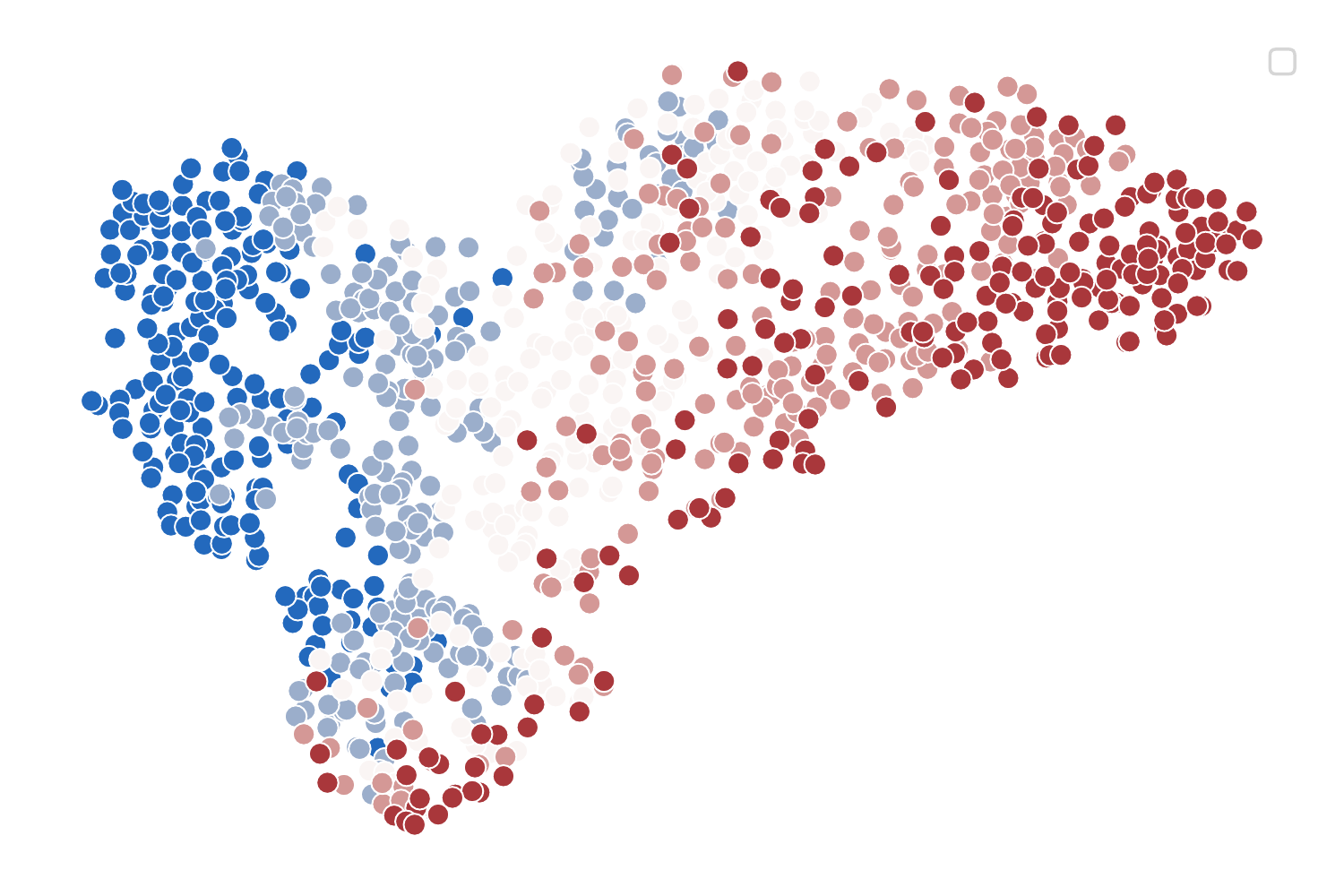}}
     \subfloat[][ia-escorts]{\includegraphics[width=.25\textwidth]{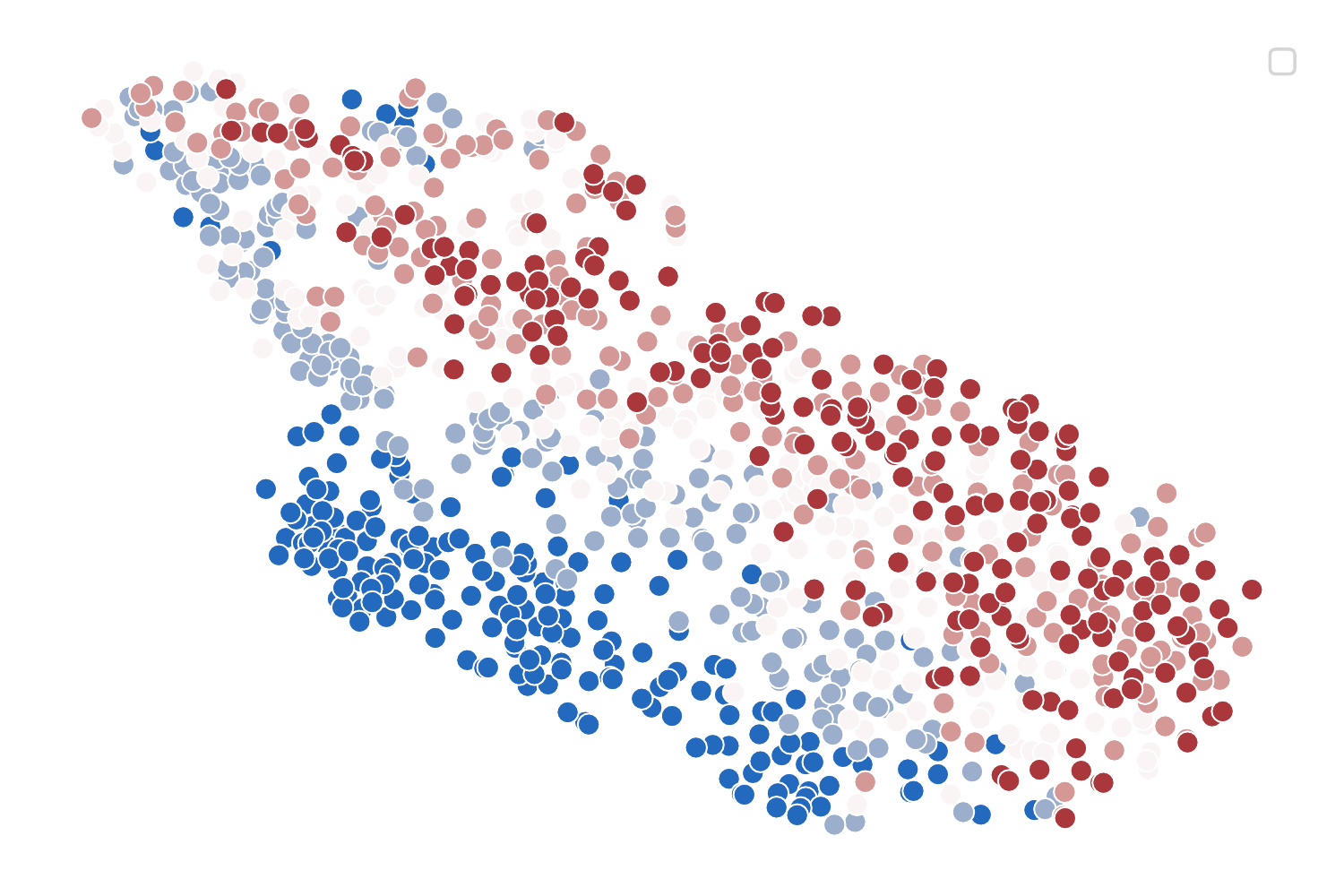}}
     \subfloat[][soc-wiki]{\includegraphics[width=.25\textwidth]{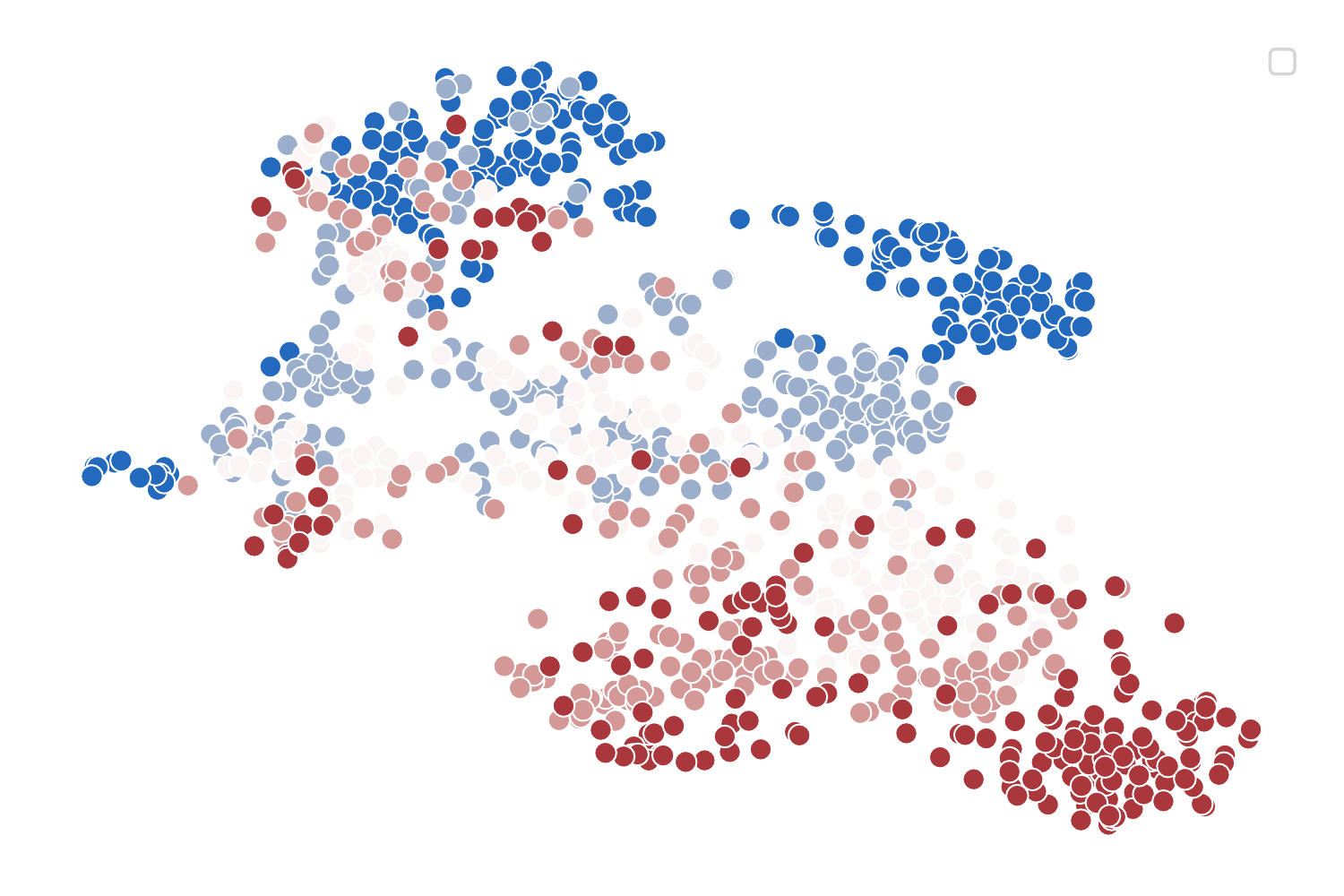}}
     \caption{The visualization results of the obtained dynamic node embeddings of different datasets. The time evolves forward {where} the color varies from dark blue and light blue to light red and dark red.}
     \label{fig:vis}
\end{figure*}

\textbf{Batch-size.}
Deep learning models update the parameters once a mini-batch using the SGD strategy.
Usually, a smaller batch size accelerates the training speed, and a larger batch size gives a more robust model.
As shown in Fig.~\ref{fig:batch_layer}(a), TAP-GNN exhibits slight changes {in} performance on most datasets under the varying batch size. 
On {the} one hand, it indicates the robustness of our proposed TAP-GNN.
On the other hand, temporal graphs used in our experiments also demonstrate stable interactions between nodes so that it is effective {in generating} embeddings from the whole temporal neighborhood.

\textbf{Number of layers.}
The number of layers plays an essential role in graph neural networks~\cite{deffe2016gcn,kipf2016semi,hamilton2017graphsage,velivckovic2017gat}.
But GNNs also suffer from the over-smoothing and overfitting problem when the number of layers is deep.
As shown in Fig.~\ref{fig:batch_layer}(b), the AUC scores of TAP-GNN on most datasets are very stable regardless of the number of layers.
Nonetheless, TAP-GNN {shows} performance degradation on the dataset \emph{ia-slash} when the number of layers is 1.
Also, baseline methods like GraphSAGE, HTNE, {and} TGAT show poor performance, as shown in Table~\ref{tab:comp}, which only access the local k-hop neighbors.
We conjecture it may be related to the dataset property as ia-slash is a directional reply dataset from the technology website.
Hence, TAP-GNN can benefit from the 2-hop neighbors largely compared to the 1-hop neighbors.

\textbf{Embedding dimensionality.}
Node embeddings aim to preserve the topology proximity in the graph with the dense vectors for each node~\cite{perozzi2014deepwalk}.
The small embedding dimensionality~\cite{dimension2018yin} may hurt the learning ability if the graph is very complicated.
The performance of TAP-GNN on fb-forum in Fig.~\ref{fig:hidden_dropout}(a) exhibits vast improvements when the embedding dimensionality keeps increasing.
Referring to other datasets, TAP-GNN performs almost invariantly, as shown in Fig.~\ref{fig:hidden_dropout}(a).

\textbf{Dropout ratio.}
The dropout strategy is a successful technique to avoid overfitting for neural networks~\cite{dropout2012hinton}, which dropouts the neurons randomly and adjusts the hidden outputs accordingly.
Our TAP-GNN also benefits from the dropout strategy slightly on most datasets, as depicted in Fig.~\ref{fig:hidden_dropout}(b).
However, it requires grid-search over the hyper-parameters to find the best dropout ratio.

\subsubsection{Visualization}

Visualization of node embeddings can assess the quality of our proposed TAP-GNN qualitatively.
We select one thousand destination nodes among all temporal edges of three datasets, obtain the dynamic node embeddings of 128-dimensions, and project them into 2D vectors using t-SNE~\cite{tsne2008maaten} as reported in Fig.~\ref{fig:vis}.
For simplicity, one thousand nodes are selected around five quantiles of the timespan, i.e., $\{0.0, 0.2, 0.4, 0.6, 0.8\}$, where each quantile contributes two hundred nodes.
The nodes are thus labeled with their corresponding quantiles and plotted with color ramps.
As shown in Fig.~\ref{fig:vis}, the time evolves forward with the color from dark blue to dark red.
It is clear that dynamic node embeddings also demonstrate distinct boundaries among different time points.
Especially, the continuous-time points exhibit proximity on the 2D plane, while the discontinuous time points like 0.0 and 0.8 exhibit clear separations visually.
In terms of the repetition ratios, the selected datasets have fewer repeated neighbors, leading to the evolving node embeddings and interaction patterns.
Otherwise, temporal graphs with high repeated ratios are similar to the static graph, which doesn't show the evolving nature as depicted in Fig.~\ref{fig:vis}.
Overall, TAP-GNN generates the dynamic node representation effectively with time evolving.

\section{Conclusion}

In this paper, we study the problem of performing message-passing on temporal graphs with respect to the varying temporal neighborhood of nodes.
A message-passing temporal graph (MPTG) is introduced to model the dynamic message-passing on temporal graphs, which also causes heavy computational complexity.
Further, we propose the temporal graph learning framework, called Temporal Aggregation and Propagation Graph Neural Networks (TAP-GNN), which updates node embeddings with their whole temporal neighbors efficiently.
Our proposed TAP-GNN outperforms other state-of-the-art baselines on temporal graphs over a variety of temporal graphs.
Specifically, TAP-GNN demonstrates robust and stable performance on extremely sparse temporal graphs.
The additional experiments on the parameter sensitivity reveal the effectiveness of different components of the proposed framework.

Nevertheless, there are still limitations of existing approaches {of} temporal graphs.
Node dynamics on temporal graphs may be caused by permanent reasons (e.g., job changes) or transient reasons (e.g., festivals), which lead to different influences on their future interactions.
For example, the job change may cause persistent influence on the email networks, while the festivals usually bring a surge in sales on e-commerce networks.
Our future direction is to develop an adaptive model to infer the latent reasons for node dynamics.

\section{Acknowledgment}

This work is partially supported by National Key R\&D Program of China under Grant No. 2022YFB2703100, the Starry Night Science Fund of Zhejiang University Shanghai Institute for Advanced Study (Grant No. SN-ZJU-SIAS-001), and the National Research Foundation Singapore under its AI Singapore Programme(Award Number:AISG2-RP-2021-023).

\ifCLASSOPTIONcaptionsoff
  \newpage
\fi

\bibliographystyle{IEEEtran}
\bibliography{ref}

\newpage
\begin{IEEEbiography}
	[{\includegraphics[width=1in,height=1.25in,clip,keepaspectratio]{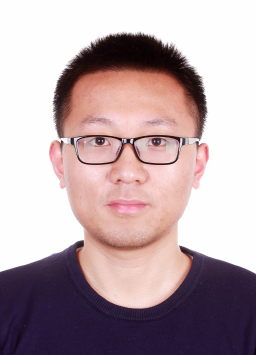}}]
	{Tongya Zheng}
	is currently pursuing the Ph.D. degree with the College of Computer Science, Zhejiang University. He received his B.Eng. Degree from Nanjing University of Science and Technology. His research interests include graph neural networks, temporal graphs, and explanation for artificial intelligence. He has authored and co-authored many scientific articles at top venues including IEEE TNNLS and AAAI. He has served with international conferences including AAAI and ECML-PKDD, and international journals including Information Sciences.
\end{IEEEbiography}
\vskip -2\baselineskip plus -1fil

\begin{IEEEbiography}
	[{\includegraphics[width=1in,height=1.25in,clip,keepaspectratio]{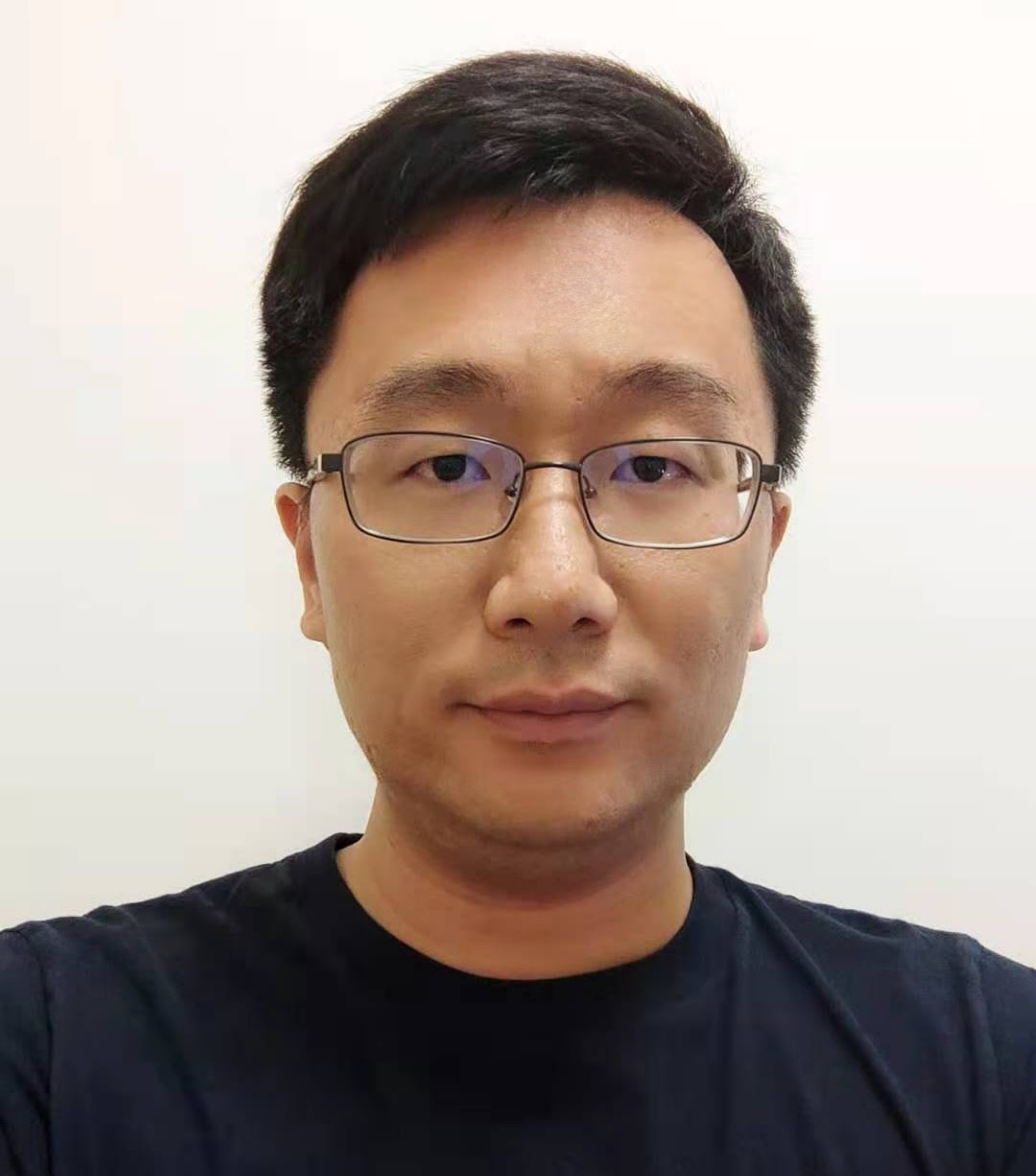}}]
	{Xinchao Wang}
	is currently an assistant professor	in the Department of Electrical and Computer Engineering, National University of Singapore (NUS). Before joining NUS, he was a tenure-track assistant professor of Computer Science at Stevens Institute of Technology, and a SwissNSF postdoctoral fellow at University of Illinois Urbana-Champaign with Prof. Thomas S. Huang. He received a PhD from Ecole Polytechnique Federale de Lausanne, and a first-class honorable degree from Hong Kong Polytechnic University. His research interests include artificial intelligence, computer vision, machine learning, medical image analysis, and multimedia. His articles have been published in major venues including CVPR, ICCV, ECCV, NeurIPS, ICLR, AAAI, IJCAI, ACL, MICCAI, TPAMI, IJCV, TIP, TKDE, and TMI. He has been or is currently serving as an associate editor of IEEE Transactions on Circuits and Systems for Video Technology,	Journal of Visual Communication and Image Representation, and Patter Recognition. He regularly serves as an area chair of CVPR, ICCV, ECCV,	NeurIPS, ICPR, ICIP, ICME, and as a senior program committee member	of AAAI and IJCAI. His team has been winner of several international challenging, including NTIRE Super-resolution Challenge’18 and AI City	Challenge’17. He received the best editor award of JVCI for 2020.
\end{IEEEbiography}
\vskip -2\baselineskip plus -1fil

\begin{IEEEbiography}
	[{\includegraphics[width=1in,height=1.25in,clip,keepaspectratio]{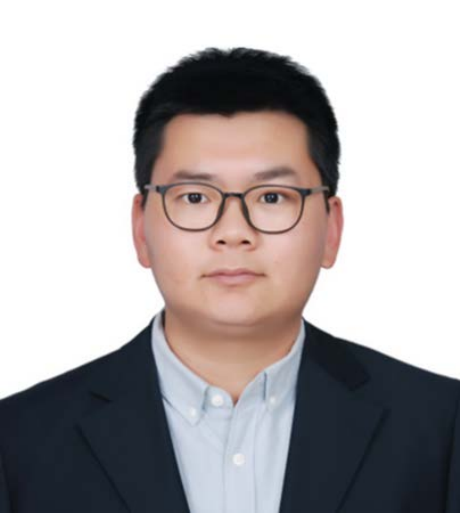}}]
	{Zunlei Feng}
	is an associate professor fellow in College of Software Technology, Zhejiang University. He received his Ph.D degree in Computer Science and Technology from College of Computer Science, Zhejiang University, and B. Eng. Degree from Soochow University. His research interests mainly include computer vision, image information processing, representation learning, medical image analysis. He has authored and co-authored many scientific articles at top venues including IJCV, NeurIPS, AAAI, TVCG, ACM TOMM, and ECCV. He has served with international conferences including AAAI and PKDD, and international journals including IEEE Transactions on Circuits and Systems for Video Technology, Information Sciences, Neurocomputing, Journal of Visual Communication and Image Representation and Neural Processing Letters.
\end{IEEEbiography}
\vskip -2\baselineskip plus -1fil

\begin{IEEEbiography}
	[{\includegraphics[width=1in,height=1.25in,clip,keepaspectratio]{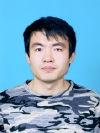}}]
	{Jie Song}
	is an assistant research fellow in College of Software Technology, Zhejiang University. He received his Ph.D degree in Computer Science and Technology from College of Computer Science, Zhejiang University, and B. Eng. Degree from Sichuan University. His research interests mainly include transfer learning, few-shot learning, model compression and interpretable machine learning.
\end{IEEEbiography}
\vskip -2\baselineskip plus -1fil

\begin{IEEEbiography}
	[{\includegraphics[width=1in,height=1.25in,clip,keepaspectratio]{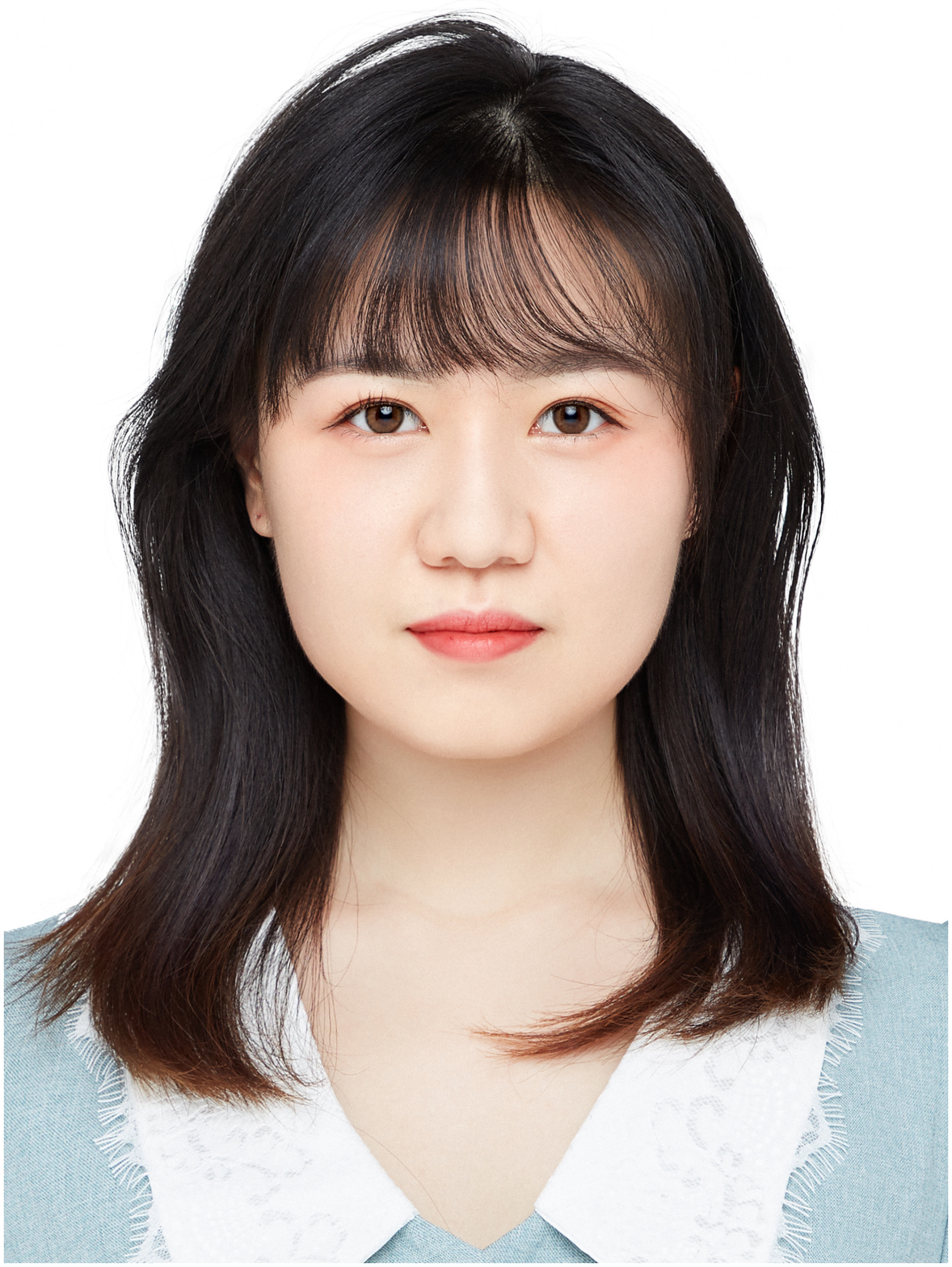}}]
	{Yunzhi Hao}
	is currently pursuing the Ph.D. degree with the College of Computer Science, Zhejiang University. Her research interests include graph inference and network embedding.
\end{IEEEbiography}
\vskip -2\baselineskip plus -1fil

\begin{IEEEbiography}
	[{\includegraphics[width=1in,height=1.25in,clip,keepaspectratio]{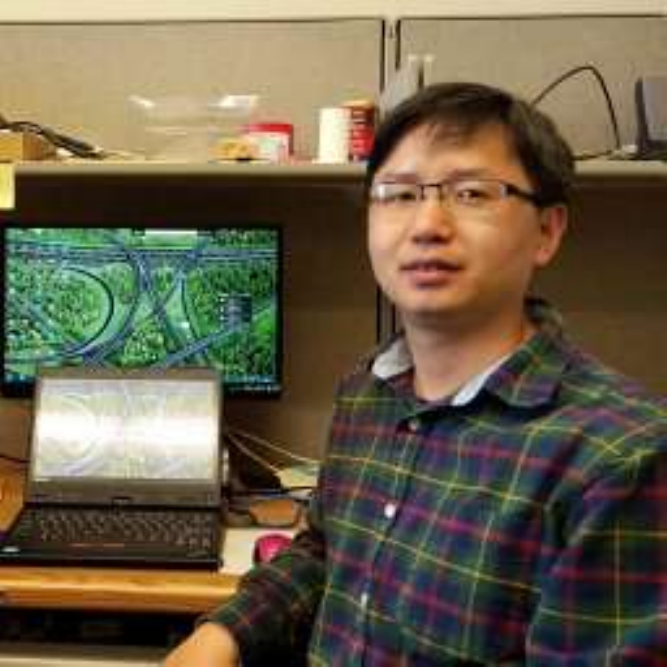}}]
	{Mingli Song}
	received the Ph.D. degree in computer science from Zhejiang University, China, in 2006. He is currently a Professor with the Microsoft Visual Perception Lab- oratory, Zhejiang University. His research interests include face modeling and facial expression analysis. He received the Microsoft Research Fellowship in 2004.
\end{IEEEbiography}
\vskip -2\baselineskip plus -1fil

\begin{IEEEbiography}
	[{\includegraphics[width=1in,height=1.25in,clip,keepaspectratio]{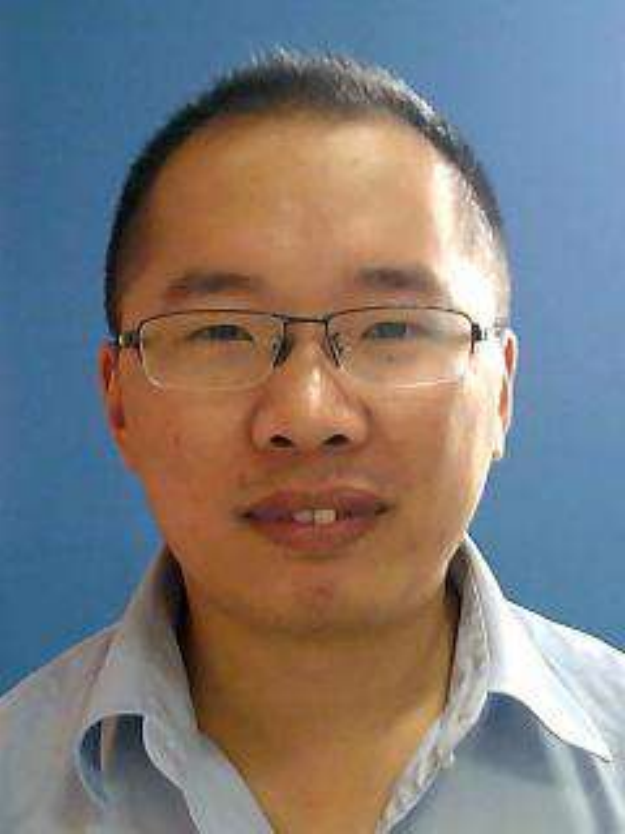}}]
	{Xingen Wang}
	received the graduate and PhD degrees in computer science from Zhejiang University of China, in 2005 and 2013, respectively. He is currently a research assistant in the College of Computer Science, Zhejiang University. His research interests include distributed computing and software performance.
\end{IEEEbiography}
\vskip -2\baselineskip plus -1fil

\begin{IEEEbiography}
	[{\includegraphics[width=1in,height=1.25in,clip,keepaspectratio]{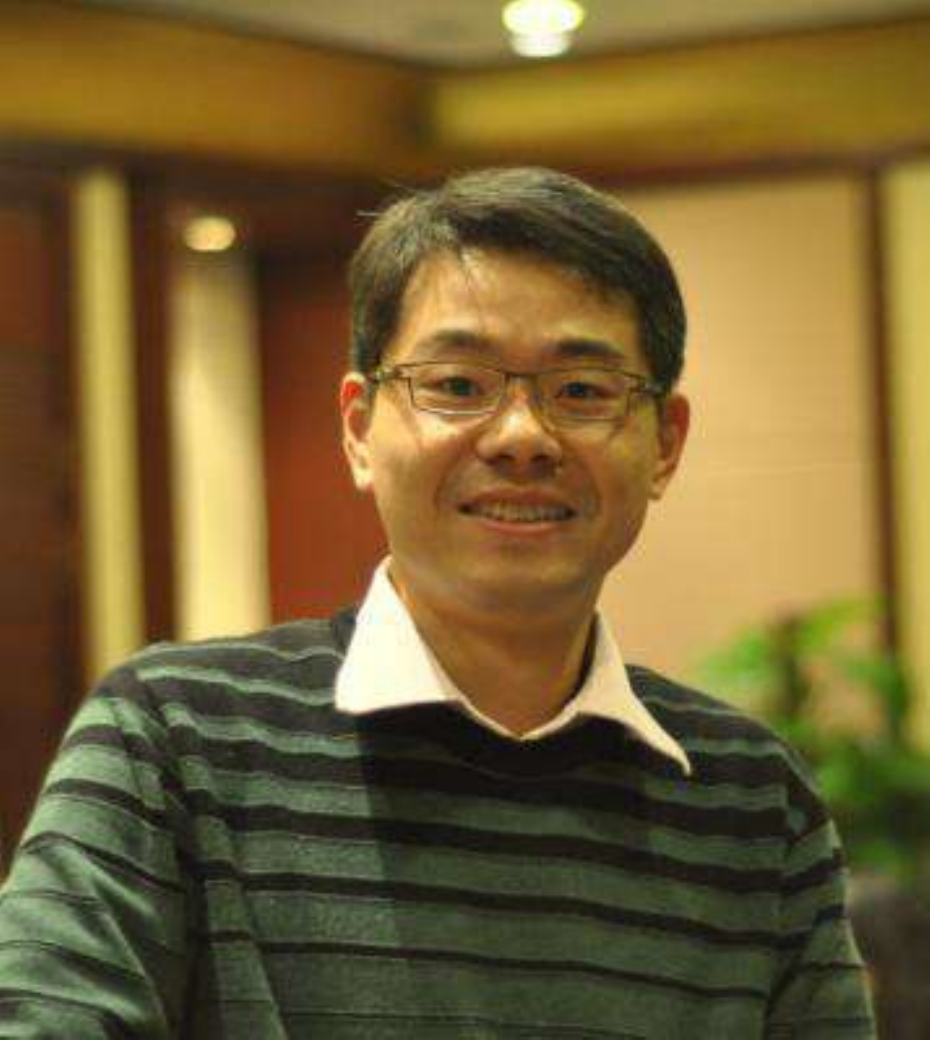}}]
	{Xinyu Wang}
	received the graduate and PhD degrees in computer science from Zhejiang University of China, in 2002 and 2007, respectively. He was a research assistant at the Zhejiang University, from 2002 to 2007. He is currently a professor in the College of Computer Science, Zhejiang University. His research interests include streaming data analysis, formal methods, very large information systems, and software engineering.
\end{IEEEbiography}
\vskip -2\baselineskip plus -1fil

\begin{IEEEbiography}
	[{\includegraphics[width=1in,height=1.25in,clip,keepaspectratio]{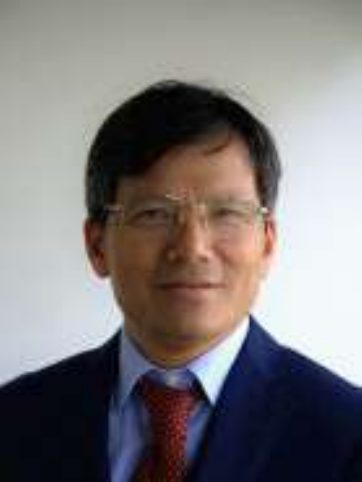}}]
	{Chun Chen}
	is currently a Professor with the College of Computer Science, Zhejiang University. His research interests include computer vision, computer graphics, and embedded technology.
\end{IEEEbiography}

\newpage
\appendix

\subsection{Proofs}
\begin{repclaim}{eq:node-size}
	The size of temporal nodes $\vert V^T \vert$ is $\mathcal{O}(\vert E \vert)$, where $E$ represents the edges in the original temporal graph.

\end{repclaim}
\begin{proof}
	Let $d_v$ denote the degree of node $v$, and we have $2 \vert E \vert = \sum_v d_v$ on graphs. $\vert V^T \vert = \sum_{v \in V} \sum_{t_k \in T_K} \mathbbm{1}\{ v_{t_k} \in V^T \} \le \sum_{v \in V} d_v = \mathcal{O} (\vert E \vert)$.
\end{proof}

\begin{repclaim}{eq:neighbor-size}
	The size of directional links $\vert E^T \vert$ defined in $G_{MP}$ is $\mathcal{O}(\sum_{v \in V} d_v^2)$, where $V$ is the original node set and $d_v$ denotes the degree of node $v$.

\end{repclaim}
\begin{proof}
	Also, let $d_v$ denote the degree of node $v$. We take a single node $v$ into account at first. Let $V_{v}^T$ be the set of temporal nodes $\{v_{t_k} | v_{t_k} \in V^T\}$ and $E_{v}^T$ be the set of directional links $\{e_{v_{t_k}} | (u, v_{t_k}) \in E^T, u \in \mathcal{TN}(v_{t_k}) \}$. Since $v_{t_i} \notin V^T$ if there are no interactions of $v$ at $t_i$, then $\vert V_v^T \vert \le d_v$. Further, as the indegree of each temporal node $\vert \mathcal{TN}(v_{t_k}) \vert \le d_v$, we have $\vert E_v^T \vert = \sum_{v_{t_k}} \vert \mathcal{TN}(v_{t_k}) \vert \le \sum_{v_{t_k}} d_v \le d_v^2$. Summing over all nodes, we have $\vert E^T \vert = \sum_{v \in V} \vert E^T_{v} \vert = \mathcal{O}(\sum_{v \in V} d_v^2)$.
\end{proof}

\subsection{{The attention kernel and its decomposition}}
\label{appendix:attention}

{
The attention mechanism is extended as a graph attention kernel by GAT~\cite{velivckovic2017gat} to impose adaptive neighborhood weights for GNNs.
Given a node $v$ and its neighbor set $\mathcal{N}(v)$, the attention kernel of GAT~\cite{velivckovic2017gat} can be written as
\begin{equation}
	\begin{split}
		\alpha_{uv} &= \text{LeakyReLU}(q^\intercal_u h_u + q^\intercal_v h_v), \\
		h_v &= \sum_{u \in \mathcal{N}(v)} \frac{\exp(\alpha_{uv}) h_u}{\sum_{u'} \exp(\alpha_{u'v})}, \\
	\end{split}
\end{equation}
where $\alpha_{uv}$ encodes the neighbor importances.
In temporal graphs, temporal neighbors also impose different importances for dynamic representation.
Let $v_{t_i}, v_{t_j} (i > j)$ be the consecutive active time points of node $v$, $u \in \mathcal{DN}(v_{t_i})$ is a neighbor emerging at $t_i$, $a(v_{t_i})$ be the embeddings of AGG, and $h(v_{t_i})$ be the embeddings of PROP.
We also need to record the cumulative attention scores $attn(v_{t_j})$ to normalize the node embeddings.
The decomposed Aggregation step is written as
\begin{equation}
	\begin{split}
		\alpha_{uv} &= \text{LeakyReLU}(q_u^\intercal h(u_{t_i}) + q_v^\intercal h(v_{t_i})), \\
		a(v_{t_i}) &= \sum_{u \in \mathcal{DN}(v_{t_i})} \exp(\alpha_{uv}) h(u_{t_i}), \\
	\end{split}
\end{equation}
where $\alpha_{uv}$ shows the importance of temporal neighbors.
The decomposed Propagation step is written as
\begin{equation}
	\begin{split}
		attn(v_{t_i}) &= attn(v_{t_j}) + \sum_{u \in \mathcal{DN}(v_{t_i})} \exp(\alpha_{uv}), \\
		h(v_{t_i}) &= a(v_{t_i}) / attn(v_{t_i}), \\
	\end{split}
\end{equation}
where $attn(v_{t_j})$ records the cumulative attention scores of last timestamp to normalize the node embeddings.
}

\subsection{{Node grouping algorithm}}
\label{appendix:node-grouping}

{
The node grouping~\cite{dgl2019wang} enables the automatic differential of computing multiple ragged lists parallelly, boosting the training speed of the PROP operation.
Firstly, we briefly review the workflow of the AP Block as shown in Algo.~\ref{alg:grouping}.
As shown in Algo.~\ref{alg:tgl}, TAP-GNN performs iterative AGG and PROP operations to obtain the high-order node embeddings for dynamic representation.
AGG works with a temporal edge set $O^T = \{ (u_{t_i}, v_{t_i}) | (u, v, t_i) \in E \}$, which connects the temporal nodes only at the same time point.
The cardinality of $O^T$ is the same as the original edge set $E$, which motivates us to send the temporal messages with $O^T$ directly.
The lined list $L^T = \{ \text{sorted}(\{ v_{t_i} | v_{t_i} \in V^T \}),  \forall v \in V \}$ organizes the temporal nodes $v_{t_i}$ referring to the same origin node $v$ into a sorted list.
The PROP operation iterates over $L^T$ to obtain the cumulative temporal node embeddings according to Section~3.4.
Secondly, we introduce the usage of the node grouping in the PROP operation.
As shown in Algo~\ref{alg:grouping}, the node grouping algorithm of DGL~\cite{dgl2019wang} works from line 6 to line 13, which parallelizes the outer loop of the PROP operation with the optimized CUDA kernels.
The inner loop from line 7 to line 12 performs a cumulative addition along with the sorted temporal nodes.
}

\begin{algorithm}[tbp]
	\caption{{Aggregation and Propagation Block}}
	\label{alg:grouping}
	\begin{flushleft}
		{\textbf{Input}: Input node embeddings $\textbf{H}(V^T)$; The temporal edge set $O^T$ and the linked list $L^T$;}
		\newline
		{\textbf{Output}: Output node embeddings $\textbf{Z}(V^T)$.}
	\end{flushleft}
	\begin{algorithmic}[1]
		\STATE {Get source and destination nodes $src, dst$ from $O^T$;}
		\STATE {// The AGG operation.}
		\STATE {$\textbf{H}_{dst} \leftarrow \textbf{H}[src]$;} 
		\STATE {$\textbf{Z}(V^T) \leftarrow \vec{0}$;}
		\STATE {// The PROP operation iterates over the linked list.}
		\FOR {{$\{v_{t_i} | v_{t_i} \in V^T\} \in L^T$}}
			\STATE {$h_v \leftarrow \vec{0}$;}
			\STATE {// Iterate over temporal nodes chronologically.}
			\FOR {{$v_{t_i} \in \{v_{t_i}\}$}} 
				\STATE {$h_v \leftarrow h_v + \textbf{H}_{dst}[v_{t_i}]$;}
				\STATE {$\textbf{Z}[v_{t_i}] \leftarrow h_v$;}
			\ENDFOR
		\ENDFOR
		\STATE {return $\textbf{Z}(V^T)$.}
	\end{algorithmic}
\end{algorithm}

\subsection{{Complexity of TAP-GNN and TGAT}}
\label{appendix:tgat}

{
We simplify the attention mechanism as a constant operation to avoid the relation of graph kernels.
The dimensions for all input features and hidden features are assumed $d$ for simplicity.
Let $N$ be the size of nodes, $M$ be the size of temporal edges, and $K$ be the number of TAP-GNN layers.
The time complexity of the TAP-GNN is $\mathcal{O}(MK(d + d^2) + Mld^2))$, where $l$ is the number of MLP layers.
Furthermore, the online time complexity of TAP-GNN is $\mathcal{O}(K(md + nd^2) + nd)$, where $m$ is the edge number of the stream graph, and $n$ is the node number of the stream graph. 
The theoretical time complexity of TGAT is reported as $\mathcal{O}(b^Kd^2)$, where $b$ is the number of sampled neighbors, $K$ is the number of TGAT layers, and $d^2$ is the transformation of node embeddings.

Compared with TAP-GNN, TGAT will suffer from the exponential expansion of sampled neighbors when increasing the number of layers.
Particularly, TAP-GNN shows much faster inference latency in the theoretical analysis.
}

{
We firstly speedup the TGAT sampler 50 times faster with the Numba\footnote{http://numba.pydata.org/} library, where the original implementation is very slow.
In Table~\ref{tab:time}, we compare TAP-GNN against TGAT with the attention kernel in Table~\ref{tab:time}, showing that TAP-GNN could achieve 3-7 times speedup with two layers and up to 68 times speedup when extended to three layers.
}

\subsection{{Discussions of The Direct Aggregation Method}}

\begin{figure}[t]
    \centering
    \includegraphics[width=\linewidth]{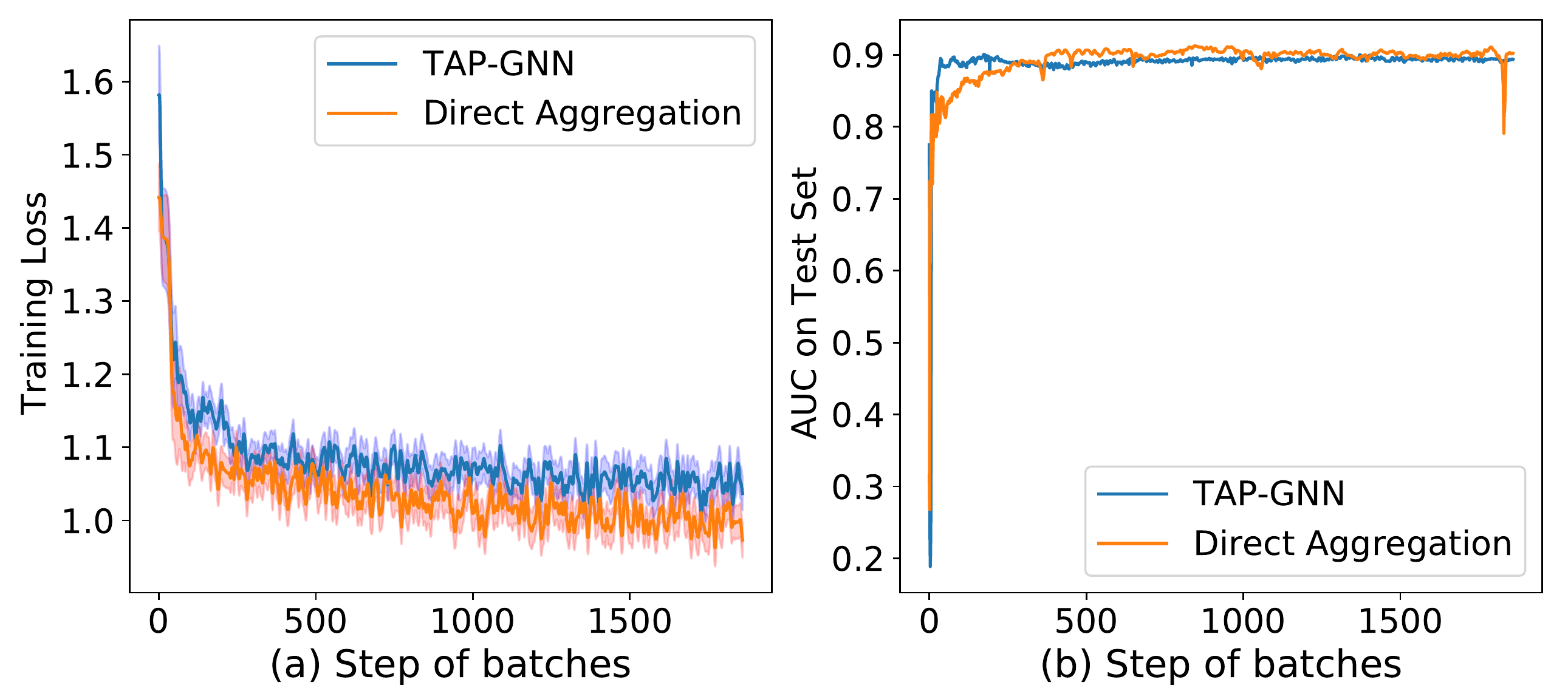}
    \caption{{Comparison of TAP-GNN against the direct aggregation method on \emph{fb-forum}. The x-axis is the number of training batches. The y-axes are the training loss and the AUC on the test set, respectively.}}
    \label{fig:direct-agg}
\end{figure}

{
Although TAP-GNN could reduce the computation complexity, the effect of the additional propagation step has not been well studied in our experiments.
We present the training progress and the AUC trend on the test set in Fig.~\ref{fig:direct-agg}.
The direct aggregation method in Fig.~\ref{fig:direct-agg} refers to a temporal GCN as defined in Eq.~\ref{eq:graph-kernel}, which costs ten times of GPU memory as our TAP-GNN.
The TAP-GNN shown in Fig.~\ref{fig:direct-agg} is implemented with GCN kernel for fairness.
Fig.~\ref{fig:direct-agg}(a) illustrates that TAP-GNN achieves a slightly larger training loss than Direct Aggregation, indicating that the propagation step may hinder the optimization of TAP-GNN.
Nonetheless, Fig.~\ref{fig:direct-agg}(b) illustrates that TAP-GNN achieves the close AUC scores more stably and faster than Direct Aggregation, indicating that the propagation step could promote the generalization of TAP-GNN on the test set.
}

\subsection{{Discussions of Discrete-Time Models}}

{
Discrete-time models~\cite{zhou2018dynamic,trivedi2017know,tnode,evolvegcn} rely on a sequence of graph snapshots to capture graph dynamics in temporal graphs.
The similarities of TAP-GNN and discrete-time models probably refer to the constructed message-passing graph, namely Aggregation-Propagation Block (AP Block).
However, similar message-passing graphs lead to different computation mechanisms and prediction performance.
}

{
Firstly, discrete-time models usually filter historical information with neural gates like LSTM, which hinders the information from long-range neighbors.
AP Block could access the whole neighborhood since it simplifies the message-passing temporal graph, as illustrated in Section~3.3.
As shown in Table~\ref{tab:comp}, discrete-time models often underperform GraphSAGE$^\star$ since the latter method could access the neighborhood directly.
Secondly, discrete-time models are designed for the full-batch update of node embeddings, while TAP-GNN updates node embeddings of a mini-batch in a graph-stream manner, as shown in Algo.~2.
The full-batch update mechanism relies on a uniform distribution of different snapshots, which suffers from the sparsity problem in a graph-stream scenario.
Thirdly, it will cause huge computation complexity when extending discrete-time models to the finest time granularity.
Let $N$ be the size of nodes, $M$ be the size of temporal edges, $S$ be the number of snapshots, and $T$ be the number of timestamps, where $T$ is usually 100 times larger than $S$.
The time complexity of one-layer discrete-time models is $\mathcal{O}(NSd^2)$, and that of TAP-GNN is $\mathcal{O}(Md^2)$.
When increasing $S$ to obtain the finest time granularity, the time complexity of discrete-time models approaches $\mathcal{O}(NTd^2)$.
Since most temporal graphs only have one interaction at each timestamp, $\mathcal{O}(NTd^2)$ will be 100 times larger than that of TAP-GNN.
Lastly, we could adopt inspiring ideas from existing discrete-time models to better mine graph dynamics.
}

\end{document}